\documentclass[twoside,11pt]{article}

\usepackage{jmlr2e}

\ShortHeadings{Why do deep convolutional networks generalize so poorly to small image transformations?}{Azulay and Weiss}
\firstpageno{1}

\usepackage{lastpage}
\jmlrheading{20}{2019}{1-\pageref{LastPage}}{6/19}{11/19}{19-519}{Aharon Azulay, Yair Weiss}
\ShortHeadings{Why do deep convolutional networks generalize so poorly to small image transformations?}{Azulay, Weiss}

\begin{document}

\title{Why do deep convolutional networks generalize so poorly to small image transformations?}

\author{\name Aharon Azulay \email aharon.azulay@mail.huji.ac.il \\
       \addr ELSC\\
       Hebrew University of Jerusalem\\
       \AND
       \name Yair Weiss \email yweiss@cs.huji.ac.il \\
       \addr CSE, ELSC\\Hebrew University of Jerusalem\\}

\editor{Rob Fergus}

\maketitle

\begin{abstract}%   <- trailing '%' for backward compatibility of .sty file

  Convolutional Neural Networks (CNNs) are commonly assumed to be
  invariant to small image transformations: either because of the
  convolutional architecture or because they were trained using data
  augmentation. Recently, several authors have shown that this is not
  the case: small translations or rescalings of the input image can
  drastically change the network's prediction. In this paper, we
  quantify this phenomena and ask why neither the convolutional
  architecture nor data augmentation are sufficient to achieve the
  desired invariance. Specifically, we show that the convolutional
  architecture does not give invariance since architectures ignore the
  classical sampling theorem, and data augmentation does not give
  invariance because the CNNs learn to be invariant to transformations
  only for images that are very similar to typical images from the training set. We discuss two possible solutions to this problem:
  (1) antialiasing the intermediate representations and (2) increasing
  data augmentation and show that they provide only a partial solution
  at best. Taken together, our results indicate that the problem of
  insuring invariance to small image transformations in neural
  networks while preserving high accuracy remains unsolved.

\end{abstract}


\section{Introduction}

\begin{figure}[ht!]
  \centering
  \includegraphics[width=0.9\textwidth]{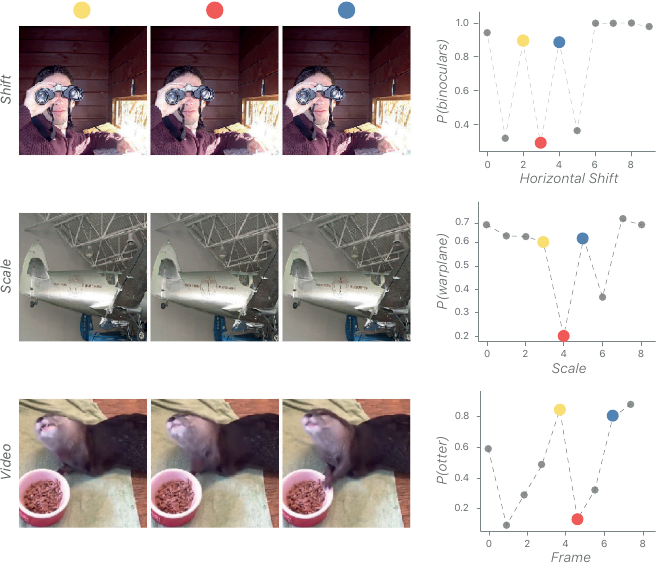}
  \caption{Examples of jagged predictions of modern deep convolutional neural networks. Top: A negligible horizontal shift of the object results in an abrupt decrease in the network's predicted score of the correct class. Middle: A tiny increase in the size of the object produces a dramatic decrease in the network's predicted score of the correct class. Bottom: A very small change in the otter’s posture results in an abrupt decrease in the network's predicted score of the correct class (see ~\url{https://youtu.be/MpUdRacvkWk}). Colored dots represent images chosen from interesting x-axis locations of the graphs on the right. These dots illustrate sensitivity of modern neural networks to small, insignificant (to a human), and realistic variations in the image.}% (see ~\url{https://youtu.be/MpUdRacvkWk}).}
  \label{fig:1}
\end{figure}

Deep convolutional neural networks (CNNs) have revolutionized computer
vision. Perhaps the most dramatic success is in the area of object
recognition, where performance is now described as ``superhuman"
\citep{he2015delving}. A key to the success of any machine learning
method is the {\em inductive bias} of the method, and clearly the
choice of architecture in a neural network significantly affects the
inductive bias. In particular, the choice of convolution and pooling
in CNNs is motivated by the desire to endow the networks with
invariance to irrelevant cues such as image translations, scalings,
and other small deformations
\citep{fukushima1982neocognitron,zeiler2014visualizing}. This
motivation was made explicit in the 1980s by Fukushima in describing
the ``neocognitron" architecture, which served as inspiration for
modern CNNs \citep{lecun1989backpropagation}. Fukushima pointed out
that the fact that all layers in the neocognitron are convolutional
means that the response in the final layer ``is not affected by the
shift in position of the stimulus pattern at all. Neither is it
affected by a slight change of the shape or the size of the stimulus
pattern.". Fukushima also demonstrated experimentally that the
neocognitron's output is unchanged when characters are translated,
rescaled or slightly deformed, even if the characters were not seen
during training.

A second source of inductive bias injected to neural networks is what is known
as ``data augmentation''. When training a CNN for object recognition,
the network is presented with a  crop of the original
image~\citep{simonyan2014very,huang2017densely}: the location and size of the crop is chosen
randomly. Thus any particular input image can be seen by the network
at different shifts and rescalings during training. We would therefore
expect that the CNN will learn a discriminant that is invariant to
resizing the image or to translating the image.

Despite these two sources of inductive bias, modern CNNs are
surprisingly brittle when the input is translated, rescaled or
otherwise transformed using a small image
transformation. Figure~\ref{fig:1} shows examples of such failures for
the InceptionResnetV2 CNN: a one pixel shift of the image (top), or a
one pixel rescaling of the image (middle), or an imperceptible change
in the otter's posture, result in a dramatic change in the network's
output. This is despite the fact that the network is fully
convolutional and was trained using data augmentation for scalings and translations.  Why does the
inductive bias fail?

In this paper we address this question. We first quantify the effect
systematically and show that it occurs in different architectures and
when different algorithms for rescaling and translating are used. We
show that the chance that a CNN output on a randomly chosen image will
change after translating downward {\em by a single pixel} can be as
high as 30\%.  We then address separately the two sources of inductive
bias (1) the convolutional architecture and (2) the use of data
augmentation and explain why they are not sufficient to achieve invariance. The convolutional architecture
ignores the classical sampling theorem, so that aliasing effects make
the output not invariant. Data augmentation causes the network to
learn invariance only for images that are very similar to those seen
during training, and since the distribution of images in the training
set is highly biased, this leads to a lack of invariance on images
that do not obey the bias.  Finally, we address two possible
solutions: antialiasing the internal representations and increasing
data augmentation and show that they provide only a partial solution,
at best.

\section{Quantifying the lack of invariance in modern CNNs}
\label{sec:quantifying_jaggedness}

\begin{figure}
%   \vspace{\floatsep}

  \begin{tabular}{@{}lrr@{}}
    \includegraphics[width=0.64\linewidth]{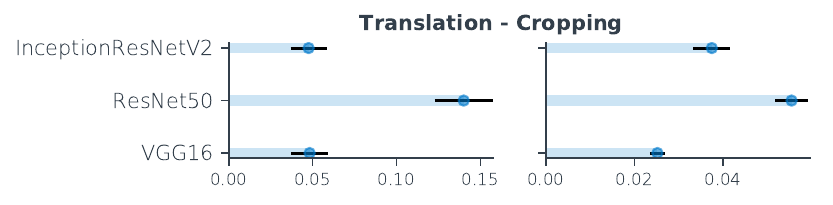} &
    \includegraphics[width=0.13\linewidth]{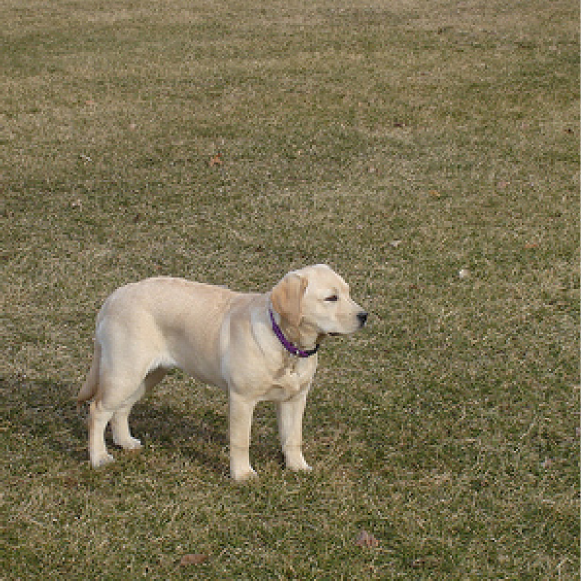} &
    \includegraphics[width=0.13\linewidth]{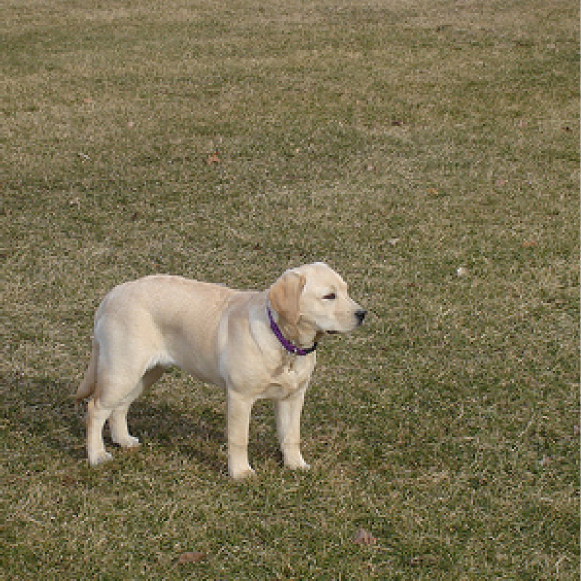}
    \\[\abovecaptionskip]
    % \small (b) Another image
  \end{tabular}
  
%   \vspace{\floatsep}

  \begin{tabular}{@{}lrr@{}}
    \includegraphics[width=0.64\linewidth]{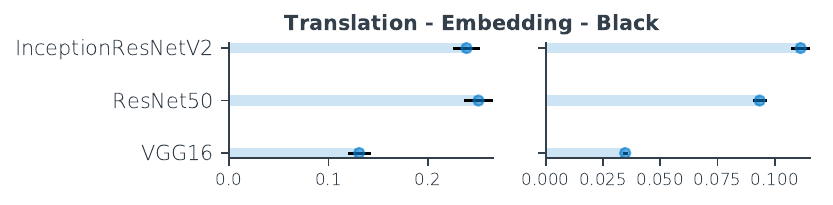} &
    \includegraphics[width=0.13\linewidth]{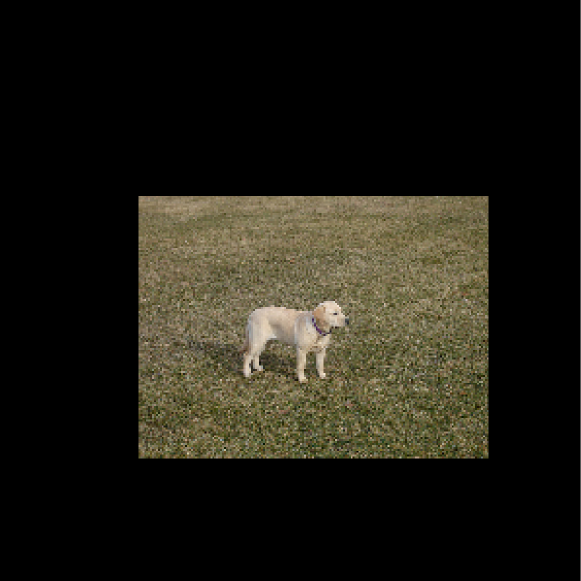} &
    \includegraphics[width=0.13\linewidth]{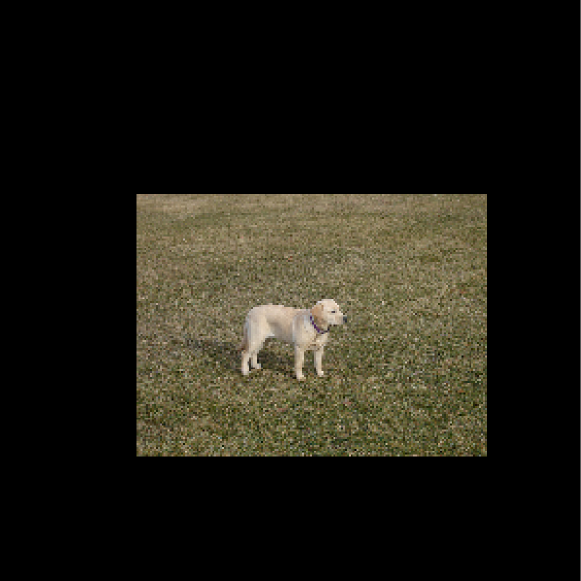}
    \\[\abovecaptionskip]
    % \small (b) Another image
  \end{tabular}
  
    \begin{tabular}{@{}lrr@{}}
    \includegraphics[width=0.64\linewidth]{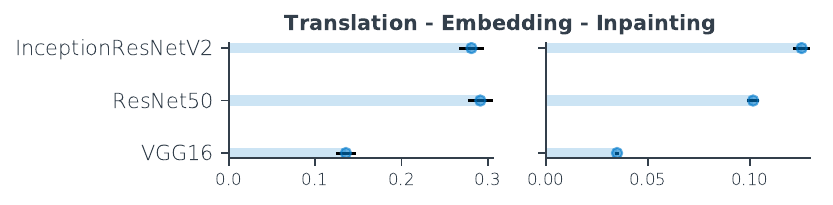} &
    \includegraphics[width=0.13\linewidth]{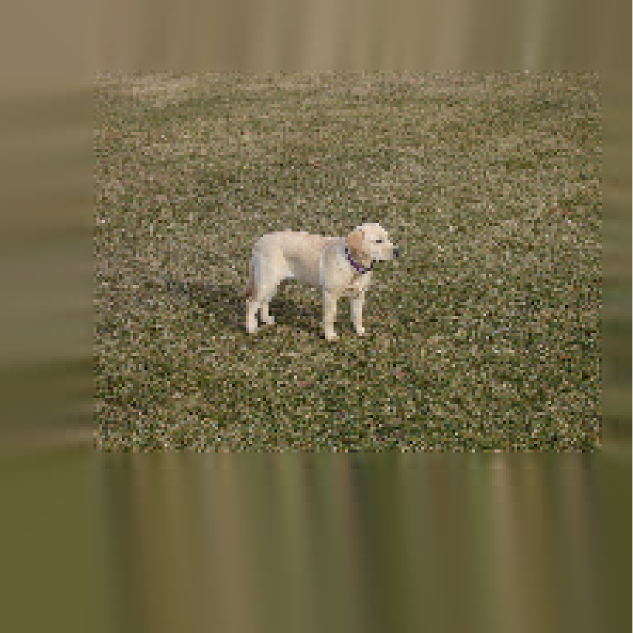} &
    \includegraphics[width=0.13\linewidth]{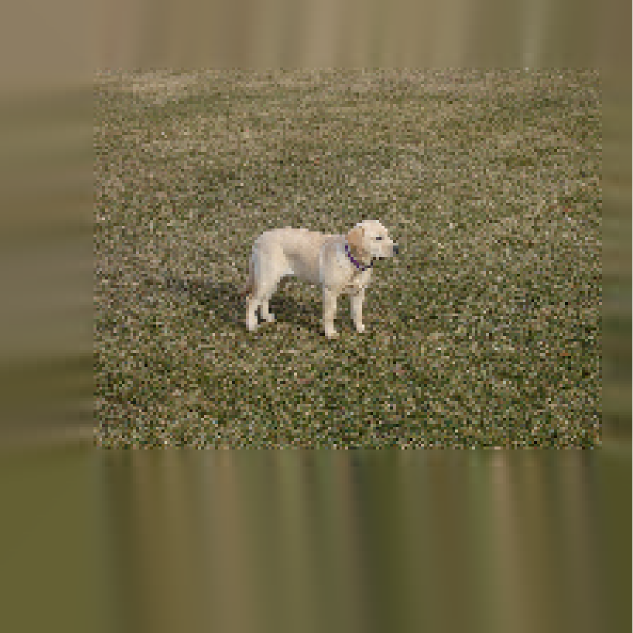}
    \\[\abovecaptionskip]
    % \small (b) Another image
  \end{tabular}
  
    \begin{tabular}{@{}lrr@{}}
    \includegraphics[width=0.64\linewidth]{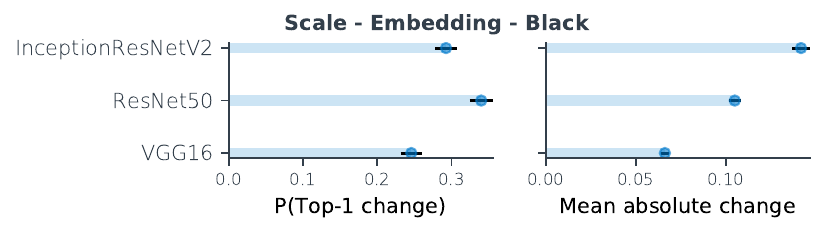} &
    \includegraphics[width=0.13\linewidth]{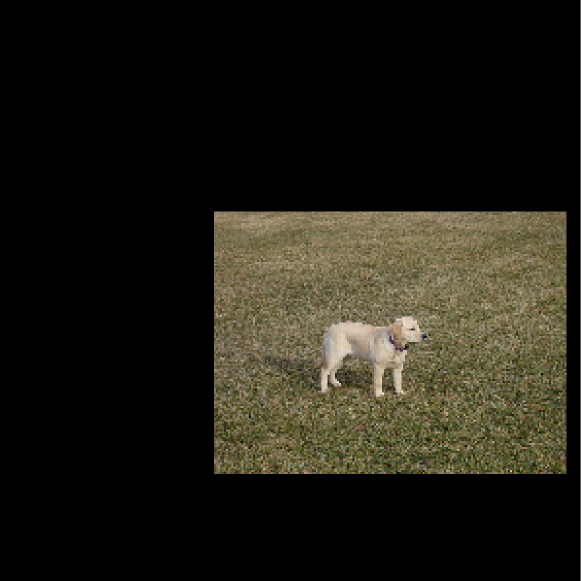} &
    \includegraphics[width=0.13\linewidth]{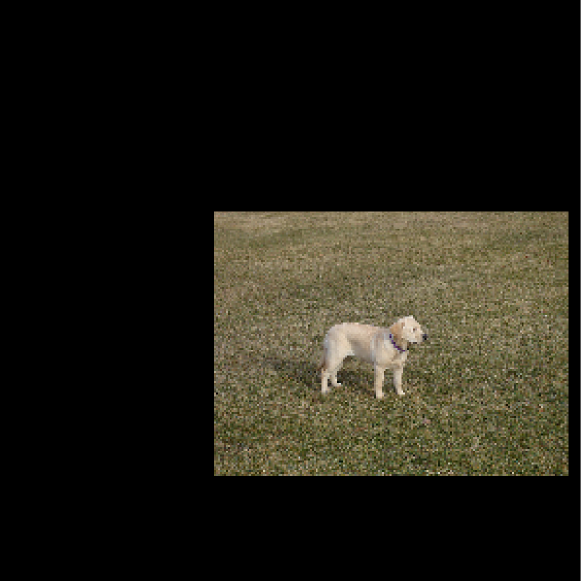}
    \\[\abovecaptionskip]
    % \small (b) Another image
  \end{tabular}
  \caption[]{Quantifying the sensitivity of modern CNNs. We tested four different protocols for one pixel perturbations (pairs of images shown on the right, animations of the transformations available at \url{https://github.com/AzulEye/CNN-Failures/blob/master/GIFS.md}) on 1000 randomly chosen images from the ImageNet validation set. For some of the protocols, the chance that a one pixel perturbation will change the output approaches $30\%$. }
  \label{fig:2}
  \end{figure}

  %  \caption{Modern deep convolutional neural networks are sensitive to small image transformations. For each of the three CNNs we tested, and for 1000 randomly chosen images from ImageNet, we tested four protocols: \textit{Top} - The standard protocol of cropping the longer side of the image to be the size of the shorter side, and calculate a measure of sensitivity for a 1-pixel translation of each image. \textit{Middle top} - We downscale the image to be 100x100 while preserving its aspect ratio, and then embed it in a 224x224 image for VGG16 and ResNet50, and 299x299 for IncetionResNetV2. We calculate a measure of sensitivity for a 1-pixel translation of each image. \textit{Middle bottom} - Same as the above protocol but with inpainted background instead of zeros background. \textit{Bottom} - Same as the above protocol but with a 1-pixel scaling instead of translation. We tested to sensitivity measures:  \textit{Left} - P(Top-1 change) is the frequency of change in the top-1 predictions following a shift. \textit{Right} - the mean absolute change in top-1 prediction following a shift. Error bars represent the standard error of the mean. Similar results for CNNs trained using a different package  are shown in the appendix. For an animated version of the transformations presented here see: \url{https://gph.is/g/EGy5XoM}
 %   
 %   }
 %    \label{fig:2}
%\end{figure}
The lack of invariance of modern CNNs to small image deformation was
reported in several recent
reports~\citep{azulay2018deep,EngstromEtal,zhang2019making}. How typical were the
examples shown in these reports? In order to systematically quantify
the effect we used 6 different CNNs and for each network we showed
1000 images with 4 different protocols.

 We tested three networks from the Keras package
(VGG16, ResNet50, InceptionResNetV2), and another three from the Pytorch
package (VGG16, ResNet50, DenseNet121). The basic experiment we conducted is
to choose a random image from the ImageNet test set, choose a one
pixel perturbation using one of the four protocols described below,
and then measure the network's sensitivity to that perturbation. We
measured the sensitivity using two measures.  The first, which we call
``P(Top-1 change)'' is the probability of change in the network's
top-1 prediction following a one pixel perturbation. The second, which we call
``mean absolute change'' measures the average absolute change in the
probability calculated by the network following a one pixel perturbation for the top class  (i.e. the class that had highest probability in the first of the two frames). We use both
measures throughout the paper since they capture different notions of
stability: the first measure is invariant to any monotonic
transformation of the output of final layer of the network, while the second
measure helps us rule out the possibility that changes in the top-1 prediction
are due to very small differences between the most likely class and
the second most likely class.

In order to perturb the input image by a single pixel, we tested four
different protocols. All of these protocols are based on the fact that
modern CNNs expect their input to be of a particular size
(typically 224x224) while the actual image can be of very different
dimensions.

 In the first protocol, which we call the ``crop" protocol we follow the procedure described in~\citep{szegedy2015going} for training CNNs.  We choose a random square within the original
 image and resize the square to be 224x224. The size and location of the square are chosen randomly according to the distribution described in~\citep{szegedy2015going}.  We then shift
 that square by one pixel diagonally to create a second image that
 differs from the first one by translation by a single pixel. In the
 second protocol, which we call ``embedding",  we downscale the image so that its minimal dimension is of size 100 while
 maintaining aspect ratio, and embed it in a random location within
 the 224x224 image, while filling in the rest of the image with black
 pixels.  We then shift the embedding location by a single pixel,
 again creating two images that are identical up to a shift by a
 single pixel. In the third protocol, we repeat the embedding
 experiment but rather than filling in the rest of the image with
 black pixels we use a simple inpainting algorithm (each black pixel is replaced by a weighted average of the non black pixels in its neighborhood. Code available at https://github.com/AzulEye/CNN-Failures). The fourth protocol is identical to the second protocol,
 but rather than shifting the embedding location, we keep the
 embedding location fixed and change the size of the embedded image by
 a single pixel (e.g from size 100x100 to size 101x101 pixels).

 Each of these protocols has its advantages and disadvantages. The
 cropping protocol by definition removes visual information that was
 present in the original image and could be important for
 discrimination. Furthermore, even though the two frames differ only by a one pixel translation, each of them has slightly different amounts of information in the border pixels.  The embedding protocol  has the advantage that all the information in the original images is preserved in both frames, but may give atypical
 appearance of the border pixels (although some images in ImageNet
 already have black borders). By examining the pairs of images
 shown in figure~\ref{fig:2} we note the following two properties:
 
\begin{itemize}
\item In all four protocols, the difference between the two images is imperceptible to a human.
\item In all four protocols, the identity of the objects in the image
  are recognizable to a human to the same degree as they were in the
  original image. 
\end{itemize}

Figure~\ref{fig:2} shows the two measures of sensitivity for the three
Keras networks tested and for the four protocols (table~\ref{comparing-pytorch-table} shows that similar numbers are obtained using the networks in the PyTorch package).   Despite the fact that the
difference between the two frames is imperceptible, the chance of a
network changing its prediction  can be as high as 30\%. In other
words, the reports of lack of invariance in previous papers were not
isolated examples but can happen with high frequency. 

In order to calibrate the significance of a failure probability of 0.3, note that unlike the case of adversarial examples (e.g. ~\citep{szegedy2013intriguing,EngstromEtal}), the perturbation is fixed and the same for all images. We are not searching for a perturbation that will fool the network, but rather applying a fixed one-pixel perturbation and measuring the probability that it will fool the network. The fact that a fixed, imperceptible perturbation fools modern networks 30\% of the time illustrates their brittleness.
% Each row corresponds to an image under different translations and the color denotes the network's estimate of the probability of the correct class. Thus a row that is all light corresponds to a correct classification that is invariant to translation, while a row that is all dark corresponds to an incorrect classification that is invariant to translation. Surprisingly, many rows show abrupt transitions from light to dark, indicating that the classification changes abruptly as the object is translated. 

A natural criticism of these results is that some of our protocols (e.g. the procedure of resizing the image and then embedding it into a larger image), was something that we introduced during testing and not during the training of the networks (we actually used {\em pretrained networks} which we downloaded from the Keras and Pytorch websites). In this respect, it is helpful to distinguish between two types of ``invariant" recognition systems:

\begin{table}
 \centerline{
\begin{tabular}{|c|c|c|c|}
\hline
& \textbf{VGG16} & \textbf{ResNet50} & \textbf{DenseNet121} \\
\hline
\textbf{P(top-1 change)}\\
\hline
Crop&  0.062 (0.05) & 0.061 (0.14) & 0.068  \\
Black & 0.18  (0.11)& 0.20 (0.29) & 0.19\\
\hline
\textbf{MAC}\\
\hline
Crop&  0.038 (0.027) & 0.04 (0.058) & 0.045  \\
Black & 0.06  (0.03)& 0.086 (0.093) & 0.079\\
\hline
\end{tabular}}
\caption[]{A comparison of the failure probability in images that were trained as part of the PyTorch package and the Keras package. The numbers show P(top1 change) and mean absolute change (MAC) for three networks trained in the PyTorch package for two of the protocols. The numbers in parenthesis are the analogous numbers for the Keras package. Although the numbers vary, the failures are common across packages and architectures.}
\label{comparing-pytorch-table}
\end{table}

\begin{itemize}
    \item A fully translation invariant recognition system will give the same output to any pattern and a translated version of that pattern. An example of such a classifier is one which bases its output only on the power spectrum of the input image. 
    \item A partially translation invariant recognition system will give the same output to a pattern and a translated version of that pattern, {\em provided that pattern appeared in the training set (or was similar to a training pattern).} An example of such a classifier is an SVM that is trained with data augmentation: by construction it is invariant to translations of the training patterns, but not to translations of different patterns.
\end{itemize}
Clearly, our results indicate that modern CNNs are not fully translation invariant (using the first definition) and can give very different outputs to a pattern and a translation of that pattern by a single pixel. 
\label{section:overview}

\section{Ignoring the Sampling Theorem}
\label{sec:Ignoring_the_Sampling_Theorem}
\begin{figure}
  \centerline{
    \includegraphics[width=0.94\linewidth]{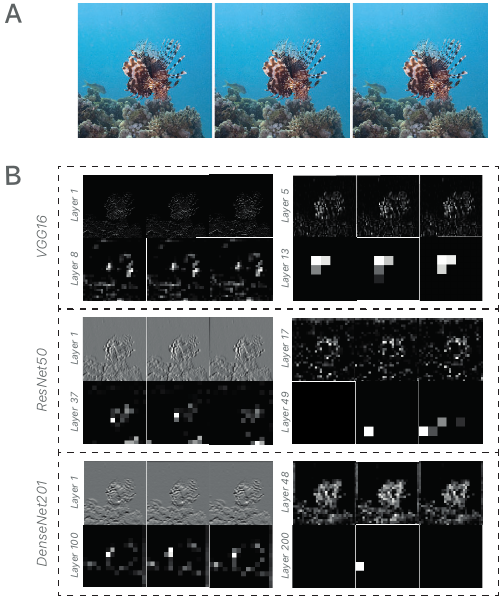} }
    \caption{The deeper the layer, the less shiftable are its feature maps. A) A vertical shift of the object in the image plane. B) Examples of feature maps from three different network architectures in response to the translated image. Layer depth assignments reflect the number of trainable convolutional layers preceding the selected layer. The last layer is always the last convolutional layer in each network.
    } \label{fig:3}
\end{figure}

The failure of CNNs to generalize to image translations is particularly puzzling. Intuitively, it would seem that if all layers in a network are convolutional then the representation should simply translate when an image is translated. If the final features for classification are obtained by a global average pooling operation on the representation (as is done for example in ResNet50 and InceptionResNetV2) then these features should be invariant to translation. Where does this intuition fail?

This intuition ignores the subsampling operation which is prevalent in modern CNNs, also known as ``stride''.
This failure of translation invariance in systems with subsampling was
explicitly discussed in~\citep{simoncelli1992shiftable} who wrote ``We cannot literally expect
translation invariance in a system based on convolution and
subsampling: translation of the input signal cannot produce simple
translations of the transform coefficients, unless the translation is
a multiple of each of the subsampling factors in the system''. 
Since deep networks often contain many subsampling operations,
the subsampling factor of the deep layers may be very large so that
``literal'' translation invariance only holds for very special
translations. In InceptionResnetV2, for example, the subsampling factor is 60, so we expect exact translation invariance to hold only for $\frac{1}{60^2}$ of possible translations. Indeed in~\cite{zhang2019making}, it was shown that invariance for modern CNNs does hold when the input pattern is shifted exactly by a multiple of the subsampling factors.

Simoncelli et al. also defined a weaker form of translation
invariance, which they called ``shiftability'' and showed that it can
hold for systems with subsampling (this is related to weak translation invariance as defined by~\citep{lenc2015understanding}, see also \citep{esteves2017polar,cohen2014transformation} for related ideas applied to neural networks). Here we extend the {basic shiftability} result to show that when shiftability holds, then global average pooling will indeed yield invariant representations.

 We define $r(x)$ as the response of a
feature detector at location $x$ in the image plane. We say that this
response is ``convolutional" if translating the image by any
translation $\delta$ yields a translation of the response by the same
$\delta$. This definition includes cases when the feature response is
obtained by convolving the input image with a fixed filter, but also
includes combinations of linear operations and nonlinear operations
that do not include any subsampling. 

We start by a trivial observation:

{\bf Observation:} If $r(x)$ is convolutional then global pooling
$r=\sum_x r(x)$ is translation invariant.

% \begin{figure}[ht!]
%  \centerline{
%  \includegraphics[width=0.5\linewidth]{Cifar_jaggedness.pdf}
%  }
%   \caption{A CNN without subsampling is perfectly translation invariant. Plotted are two Alexnet style CNNs on the CIFAR-10 dataset: with (gray) and without (red) subsampling. The two CNNs achieve similar accuracies of about 0.8.} 
%     \label{fig:CIFAR}
% \end{figure}

{\bf Proof:} This follows directly from the definition of a
convolutional response. If $r(x)$ is the feature response to one
image and $r_2(x)$ is the feature response to the same image
translated, then $\sum_x r(x)=\sum_x r_2(x)$ since the two responses
are shifts of each other.

We emphasize that the claim above guarantees that a CNN where the stride is always one, will be fully translation invariant: it will give the same output to any pattern and a translated version of that pattern, regardless of whether such a pattern is similar to one that it saw during training. We now show that this can also be achieved with subsampling, provided the representations are ``shiftable".

{\bf Definition:} A feature detector $r(x)$ with subsampling factor
$s$ is called ``shiftable'' if for any $x$ the detector output at
location $x$ can be linearly interpolated from the responses on the
sampling grid:
$$
r(x) = \sum_i B(x-x_i) r(x_i) 
$$
where $x_i$ are located on the sampling grid for subsampling factor
$s$ and $B(x)$ is the basis function for reconstructing $r(x)$ from
the samples. 

The classic Shannon-Nyquist theorem tells us that $r(x)$ will be
shiftable if and only if the sampling frequency is at least twice the
highest frequency in $r(x)$.

{\bf Claim:} If $r(x)$ is shiftable then global pooling on the
sampling grid $r=\sum_i
r(x_i)$ is translation invariant.

{\bf Proof:} This follows from the fact that global pooling on the
sampling grid is (up to a constant) the same as global pooling for all
$x$.
\begin{eqnarray}
\sum_x r(x) &=& \sum_x \sum_i r(x_i) B(x-x_i) \\
                   &=& \sum_i r(x_i) \sum_x B(x-x_i) \\
                   &=& K \sum_i r(x_i)
\end{eqnarray}
where $K=\sum_x B(x-x_i)$ and $K$ does not depend on $x_i$. 

While the claim focuses on a global translation, it can also be extended to piecewise constant transformations. 

{\bf Corollary:} Consider a set of transformations $T$ that are constant on a set of given image subareas. If $r(x)$ is shiftable and for a given image, the support of $r(x)$ and its receptive field is contained in the same subregion for all transformations in $T$, then global pooling on the sampling grid is invariant to any transformation in $T$. 

{\bf Proof:}
This follows from the fact that applying any transformation in $T$ to an image has the same effect on the feature map $r(x)$ as translating the image.

To illustrate the importance of the sampling theorem in guaranteeing invariance in CNNs, consider a convolutional layer in a deep CNN where each unit acts as a localized ``part detector" (this has been reported to be the case for many modern CNNs~\citep{zeiler2014visualizing,ZhouKLOT14}). Each such part detector has a spatial tuning function and the degree of sharpness of this tuning function will determine whether the feature map can be subsampled while preserving shiftability or not. For example, consider a part detector that fires only when the part is exactly at the center of its receptive field. If there is no subsampling, then as we translate the input image, the feature map will translate as well, and the global sum of the feature map is invariant to translation. But if we subsample by two (or equivalently use a stride of two), then there will only be activity in the feature map when the feature is centered on an even pixel, but not when it is centered on an odd pixel. This means that the global sum of the feature map will {\em not} be invariant to translation. 

In the language of Fourier transforms, the problem with a part detector that fires only when the part is exactly at the center of the receptive field is that the feature map contains many high frequencies and hence it cannot be subsampled while preserving shiftability. On the other hand, if we have a part detector whose spatial tuning function is more broad, it can be shiftable and our claim (above) shows that the global sum of activities in a feature map will be preserved for all translations, even though the individual firing rates of units will still be different when the part is centered at an odd pixel or an even pixel. Our corollary (above), shows the importance of shiftability to other smooth transformations: in this case each ``part detector" will translate with a {\em different} translation but it is still the case that nonshiftable representations will not preserve the global sum of activities as the image is transformed, while shiftable representations will.

% \begin{figure}[ht!]
% \centering
%   \begin{tabular}{@{}l@{}}
%     \includegraphics[width=0.65\linewidth]{3AspectRatio.png} \\
%     % \small (a) An image
%   \end{tabular}

% %   \vspace{\floatsep}

% %   \begin{tabular}{@{}l@{}}
% %     % \small C
% %     \includegraphics[width=0.65\linewidth]{Top1VsDepth.pdf} \\
    
% %   \end{tabular}
  
%   \caption{The deeper the network, the less shiftable are the feature maps. A) A vertical shift of a "Kuvasz" dog in the image plane. B) Examples of feature maps from three different network architectures in response to the translated Kuvasz image. Layer depth assignments reflect the number of trainable convolutional layers preceding the selected layer. The last layer is always the last convolutional layer in each network.}
%     \label{fig:3}
% \end{figure}

Figure~\ref{fig:3} examines the extent to which the representations learned by modern CNNs are invariant or shiftable. The top row shows an image that is translated vertically, while the bottom three rows show typical representations in different layers for the three CNNs we consider. For VGG16 the representations appears to shift along with the object, including the final layer where the blurred pattern of response is not a simple translation of the original response, but seems to preserve the global sum for this particular image. For the two more modern networks, the responses are sharper but lose their shiftability in the later layers. In particular, the final layers show approximate invariance to one special translation but no response at all to another translation, suggesting that the many layers of subsampling yield a final response that is not shiftable.

To gain a more quantitative measure of how shiftability changes in modern networks as a function of depth, we took the pretrained modern networks and trained a readout layer to classify ImageNet images from intermediate layers. For example, the InceptionResNetV2 network has 134 layers, but we can train a readout layer to classify images based on each of the preceding 133 layers. After training these readout layers, we measure whether a one pixel shift of the input would cause a change in the networks output which allows us to quantify the amount of translation invariance in the different layers. 

\begin{figure}
\centerline{
    \includegraphics[width=0.7\linewidth]{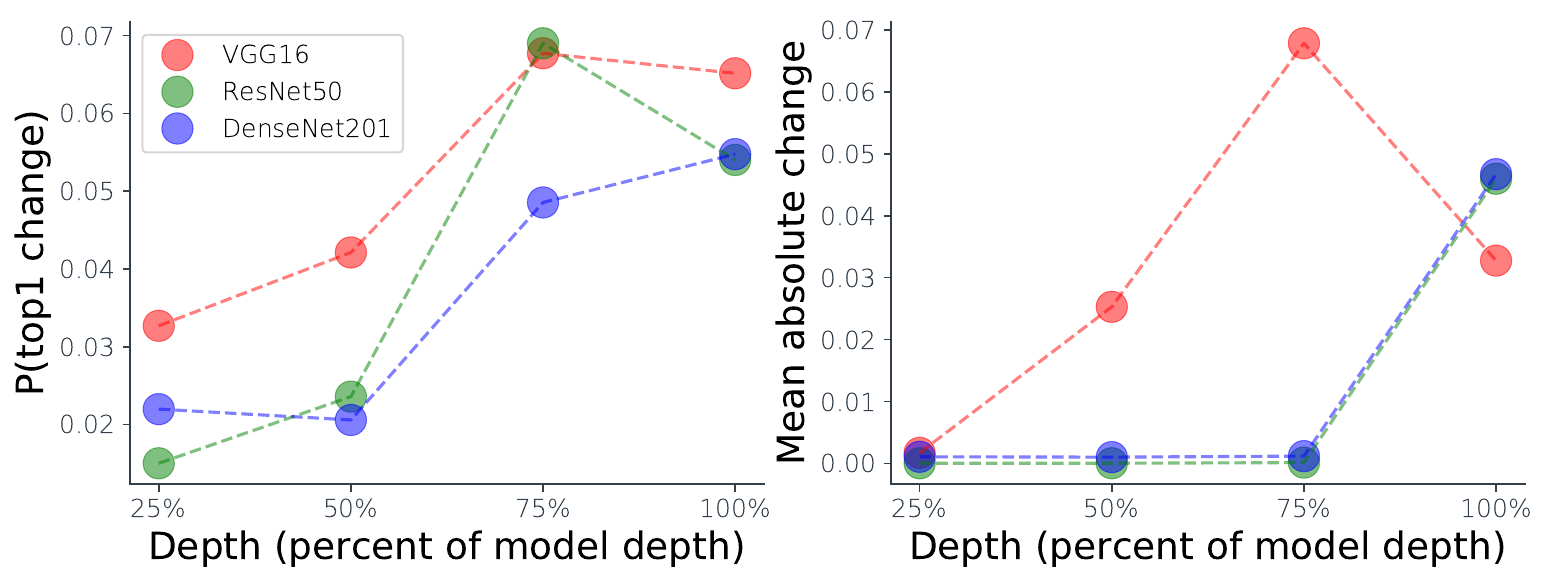}}
    \caption{The deeper the representation, the higher the probability for a change in the top-1 prediction (left) and the mean absolute change (right) following a 1-pixel translation (using the cropping protocol). Shown are trained readout layers of three different CNNs at different depths (relative to each CNN).
    } 
    \label{fig:quantitativeJaggedness}
\end{figure}

Results are shown in figure~\ref{fig:quantitativeJaggedness}: when we train classifiers based on the early layers in the input, the chance of a one pixel shift changing the output is below 5\% but as we go deeper and deeper into the network, the subsampling operations and the nonlinearities make the representations not shiftable and the network loses its invariance. 

%put here the new results with retraining intermediate representations?

How can we guarantee that representations in CNNs will be shiftable? As explained above, we need to make sure that any feature map that uses stride does not contain frequencies above the Nyquist frequency. If CNNs were purely linear, we could simply blur the input images so that they would not include any frequencies higher than the Nyquist limit determined by the final sampling factor of the network. But since CNNs also include nonlinearities, they can add high frequencies that were not present in the input.

{\bf Observation 2:} Let $r(x)$ be a representation obtained by first subsampling an image by a factor of $s$ and then applying a series of convolutions and pointwise nonliearities. Let $r_2(x)$ be a representation obtained without subsampling the image, and then applying the same series of convolutions and pointwise nonlinearities, but where all convolution kernels are dilated by a factor of $s$. Then subsampling $r_2(x)$ by a factor of $s$ yields $r(x)$.

{\bf Proof:} This follows directly from the definition of convolution.

{\bf Corollary:} Consider a layer in a CNN with stride $s$. For any {\em subsequent} layer in the CNN, $r(x)$, consider the equivalent $r_2(x)$ as defined in observation 2.  If $r_2(x)$ contains frequencies above the Nyquist frequency of $s$, then $r(x)$ is not shiftable.

{\bf Proof:} This follows from observation 2, which states that we can calculate $r(x)$ by first computing $r_2(x)$ and then subsampling by a factor of $s$.

Observation 2 and its corollary highlight an important difference between CNNs and linear systems. In a linear system, we can always avoid aliasing by {\em blurring before subsampling.} The blur removes frequencies above the Nyquist limit, so that the subsampling does not cause aliasing. Furthermore, since all subsequent operations are linear, they cannot introduce frequencies that were zeroed out by the blur, hence all subsequent representations are shiftable. But in nonlinear systems, this no longer holds. Even if we are careful to remove all frequencies above the Nyquist limit before sampling, there is no guarantee that the subsequent representations will remain shiftable. 

Translated to the language of neural networks the principle of blurring before subsampling means that stride (i.e. subsampling) should always be combined with pooling (i.e. blurring) in the preceding layer. Indeed if we have an arbitrarily deep CNN where all the layers use stride=1 followed by one layer that has a stride greater than one, then by choosing the pooling window appropriately we can guarantee that the final layer will still be shiftable. If we do not use appropriate pooling then there is no guarantee that this layer will be shiftable.  Even if we use appropriate pooling that ensures that a given layer is shiftable, the subsequent nonlinearities in a CNN may not preserve the shiftability, as the nonlinearities may again introduce high frequencies. The recent paper of~\citep{zhang2019making}  reports on experiments that add blur prior to subsampling in modern CNNs (specifically they add the blur after the ReLU operation). We will examine the extent that such antialiasing helps invariance in section~\ref{sec:antialiasing}.

\section{Why don't modern CNNs learn to be invariant from data augmentation?}
%\label{others}

\begin{figure}
\centerline{
    \begin{tabular}{cccc}
 \includegraphics[width=0.2\textwidth]{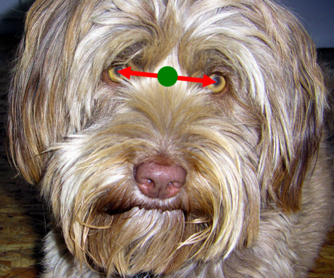} & 
 \includegraphics[width=0.2\textwidth]{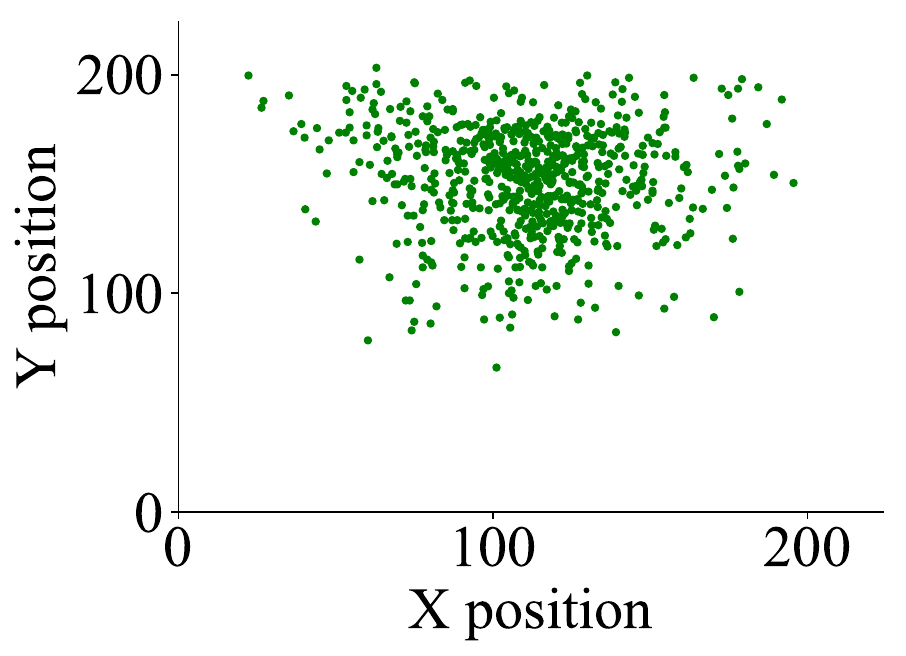} & 
 \includegraphics[width=0.2\textwidth]{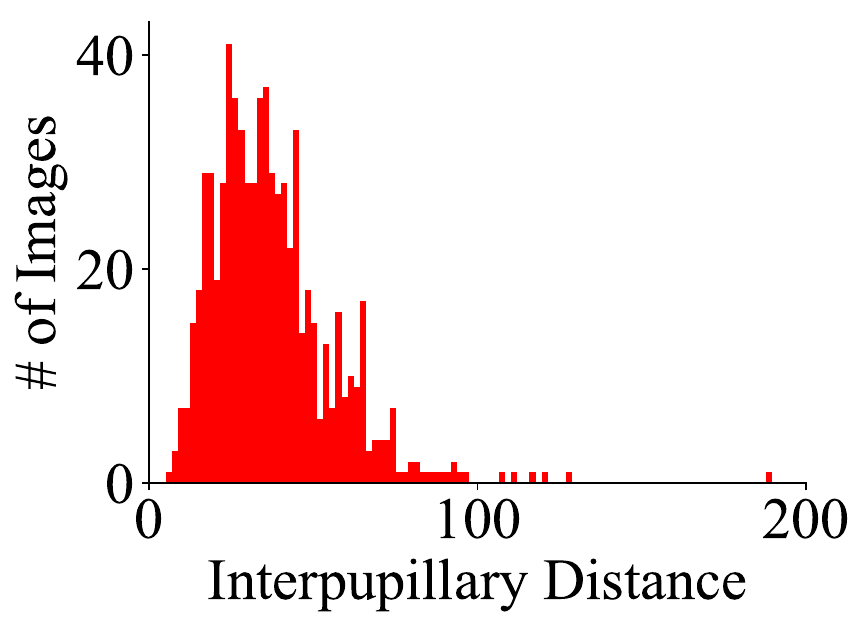} &
 \includegraphics[width=0.2\textwidth]{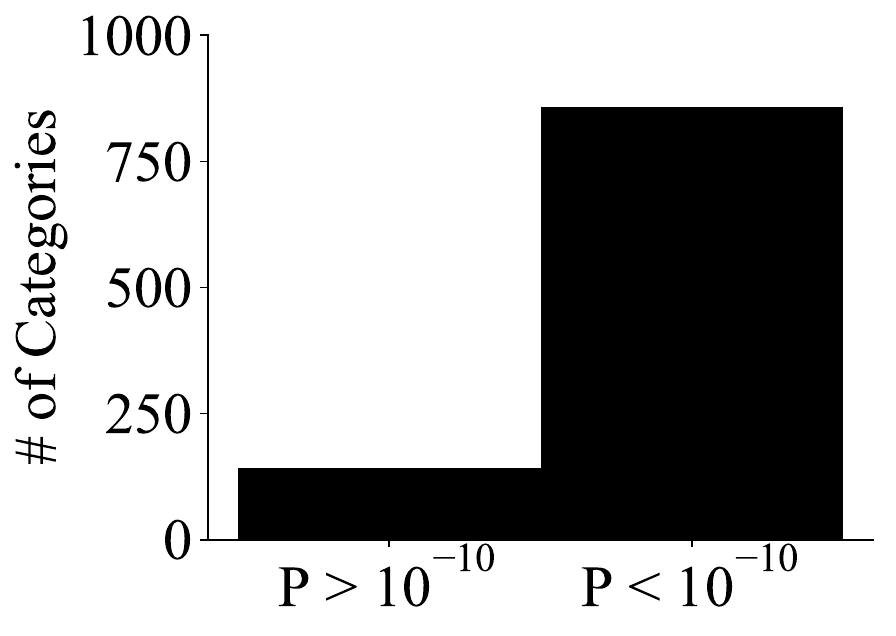} 
 \end{tabular}}
    \caption{Photographer's biases in the ImageNet dataset. Left: Example of the hand-labeling procedure of the “Tibetan terrier” category. Middle: Location and Histogram of distances between the dog’s eyes. Notice the bias in both the object’s position and scale. Right: Quantitative results for all ImageNet categories (Chi-squared test).
    } 
    \label{fig:4}
\end{figure}

\begin{figure}
\centerline{
    \includegraphics[width=0.15\linewidth]{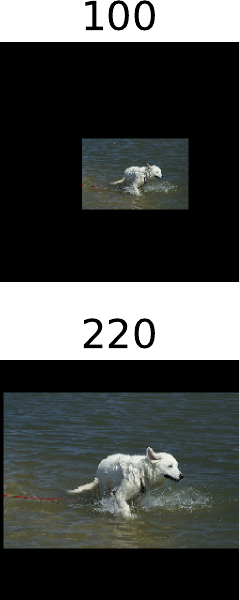} 
    \includegraphics[width=0.40\linewidth]{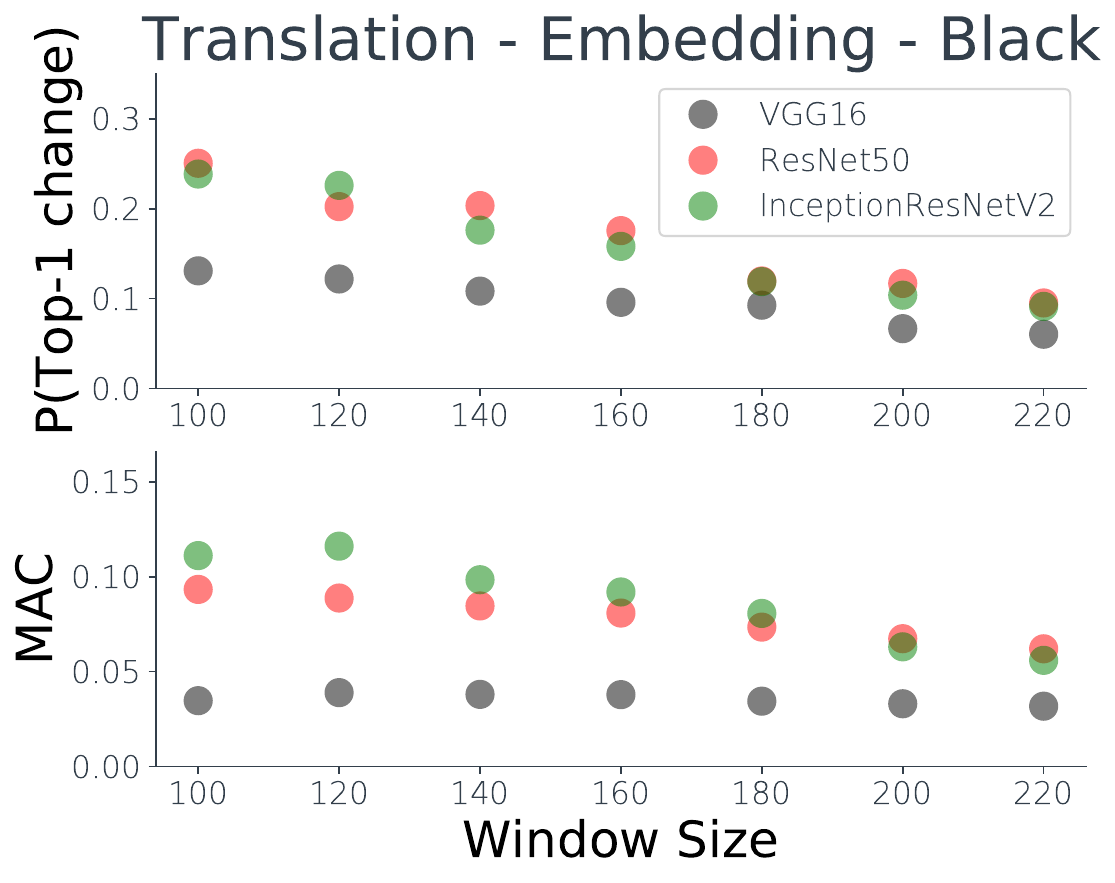} 
    \includegraphics[width=0.40\linewidth]{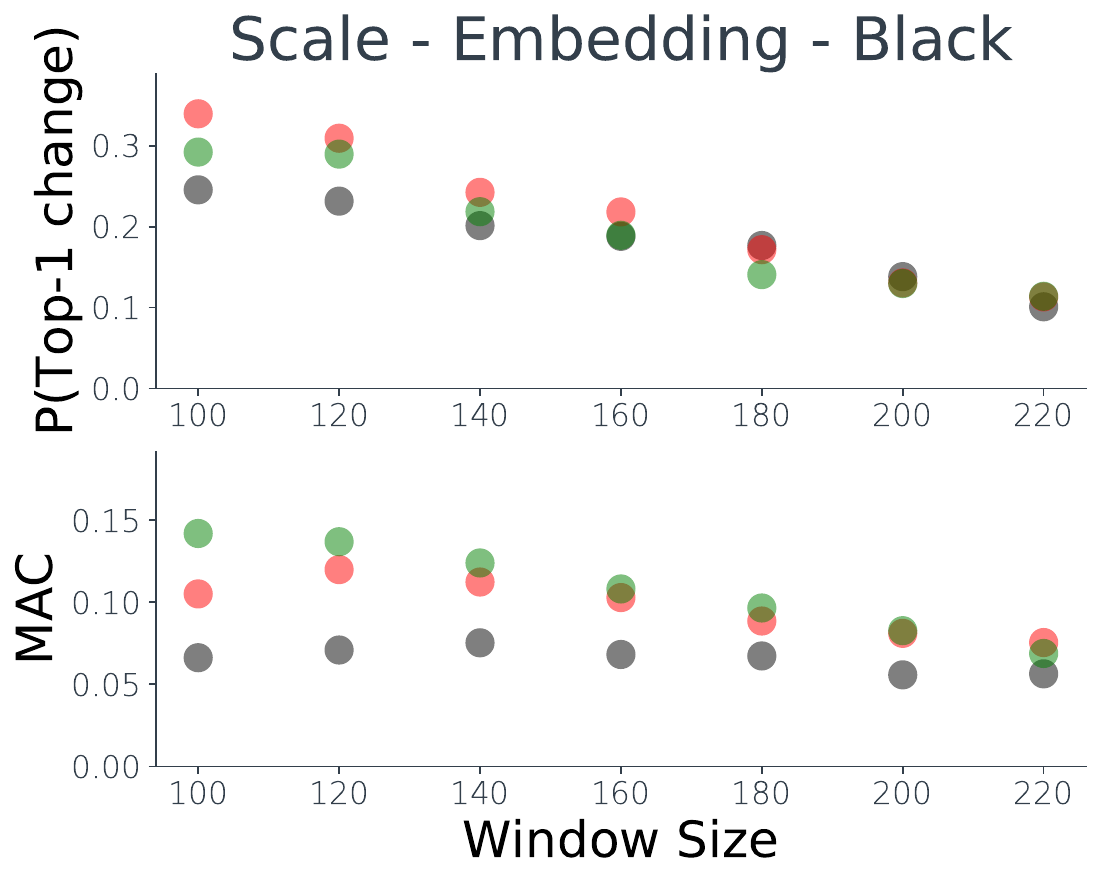}}
    \caption[]{The effect of the embedding window size on the observed failures. When the embedding size is different from the size that the network expects, the lack of invariance increases. MAC stands for the mean absolute change measure.} 
    \label{fig:6}
\end{figure}

\begin{figure}[ht!]
\centering
     \begin{tabular}{ccccc}
     
 & ~~~~~~Typical & ~~~~~Atypical \\
 \raisebox{32pt}[0pt][0pt] & \includegraphics[width=0.48\textwidth]{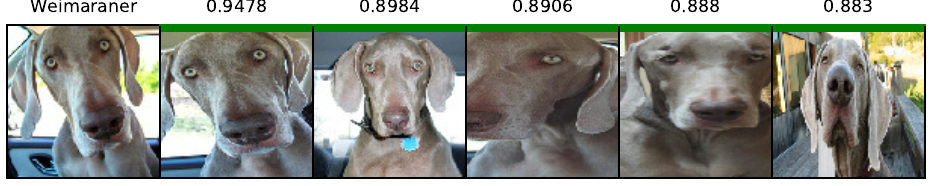} & 
 \includegraphics[width=0.48\textwidth]{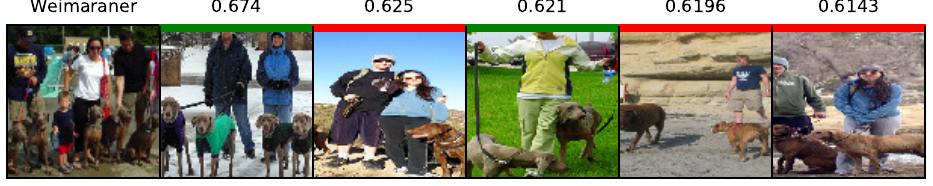} \\
 \raisebox{43pt}[0pt][0pt] &  \includegraphics[width=0.48\textwidth]{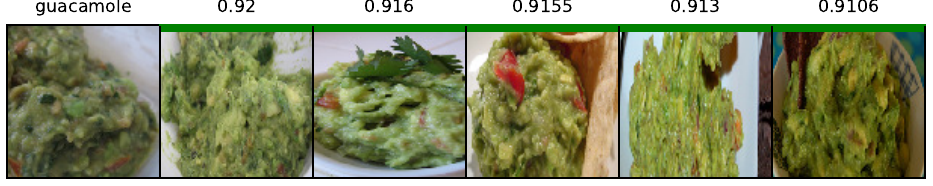} & 
 \includegraphics[width=0.48\textwidth]{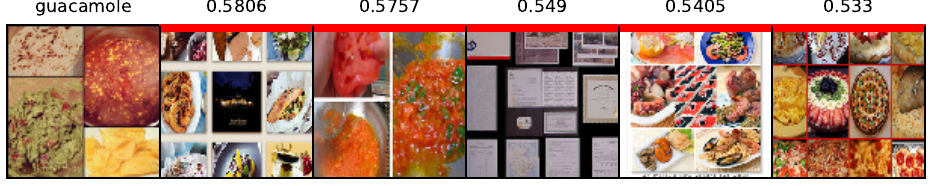} \\
 \raisebox{43pt}[0pt][0pt] &  \includegraphics[width=0.48\textwidth]{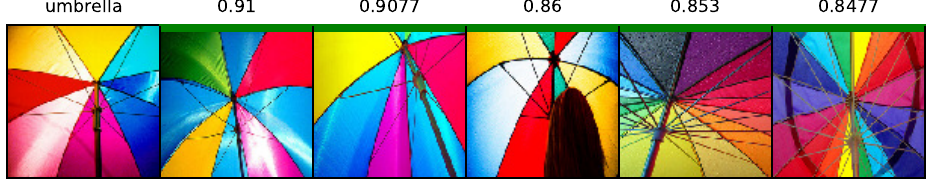} & 
 \includegraphics[width=0.48\textwidth]{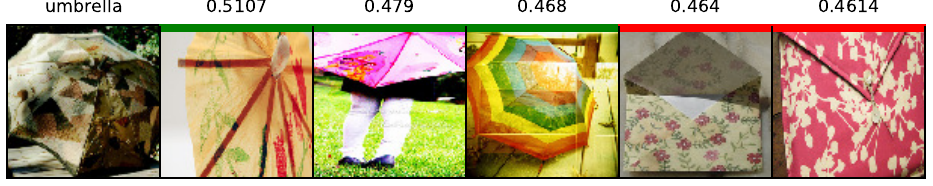} \\
 \raisebox{43pt}[0pt][0pt] &  \includegraphics[width=0.48\textwidth]{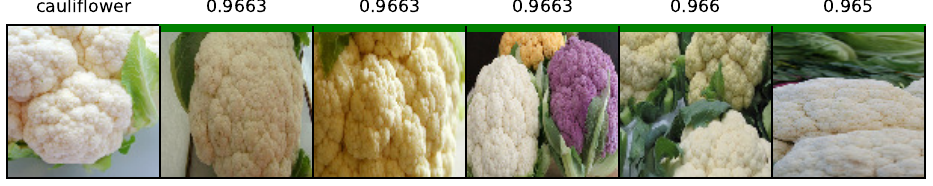} & 
 \includegraphics[width=0.48\textwidth]{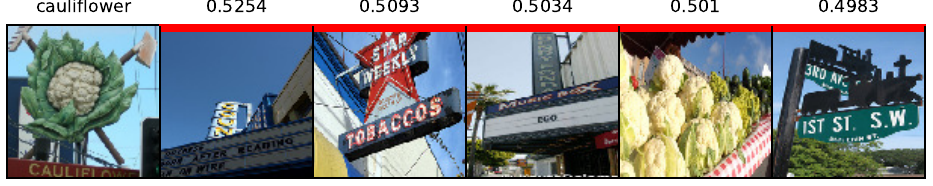} \\
 \raisebox{43pt}[0pt][0pt] &  \includegraphics[width=0.48\textwidth]{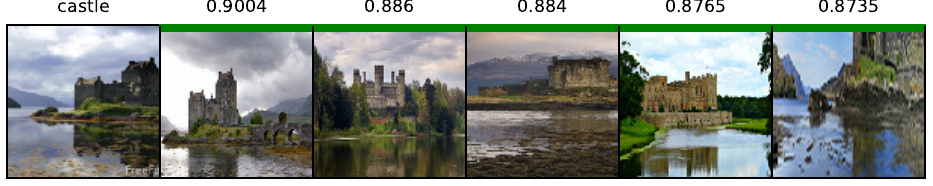} & 
 \includegraphics[width=0.48\textwidth]{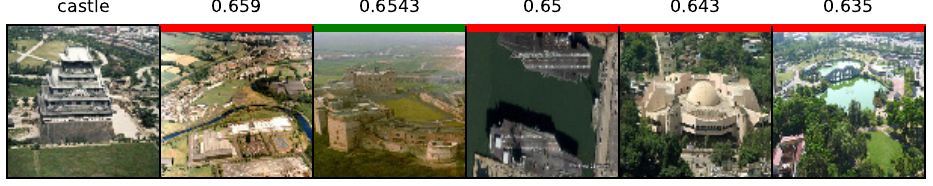} \\
 \raisebox{43pt}[0pt][0pt] &  \includegraphics[width=0.48\textwidth]{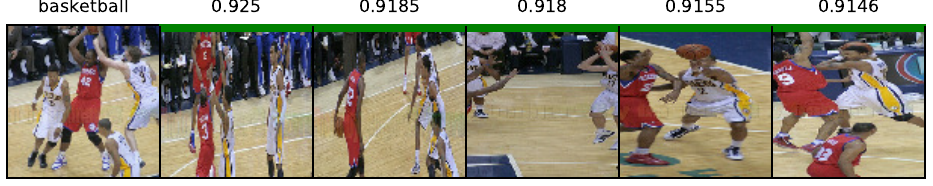} & 
 \includegraphics[width=0.48\textwidth]{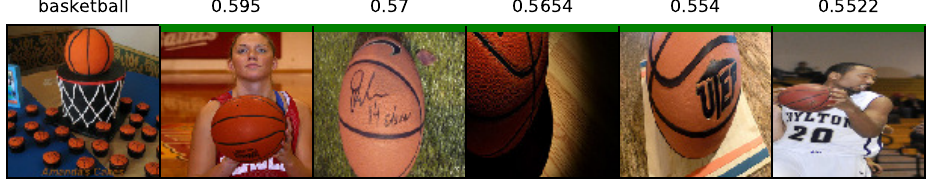} 
 \end{tabular}
    \caption{Our definition of typical images (i.e. ones that satisfy photographer's bias). Examples of randomly chosen, center cropped validation images, for which we calculate the normalized correlation of representation in the global-average-pooling layer of ResNet50, to the same representation of the entire ImageNet training set.
    Left: Images with a high mean correlation for their top 10 nearest neighbors from the training set (top 5 NN are shown on the right of each validation image, together with their correlation to the validation image).
    Right: The same as the left side, but with images with a low mean correlation for their top 10 nearest neighbors from the training set.
    } \label{fig:typicality}
\end{figure}

\begin{figure}
  \centering
  \begin{tabular}{@{}c@{}}
    \includegraphics[width=.8\linewidth]{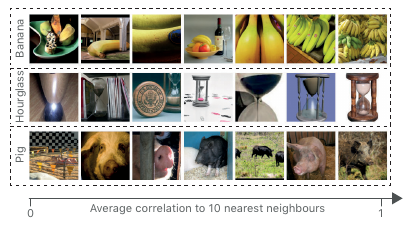} \\[\abovecaptionskip]
  \end{tabular}

  \begin{tabular}{@{}c@{}@{}c@{}}
    \includegraphics[width=.4\linewidth]{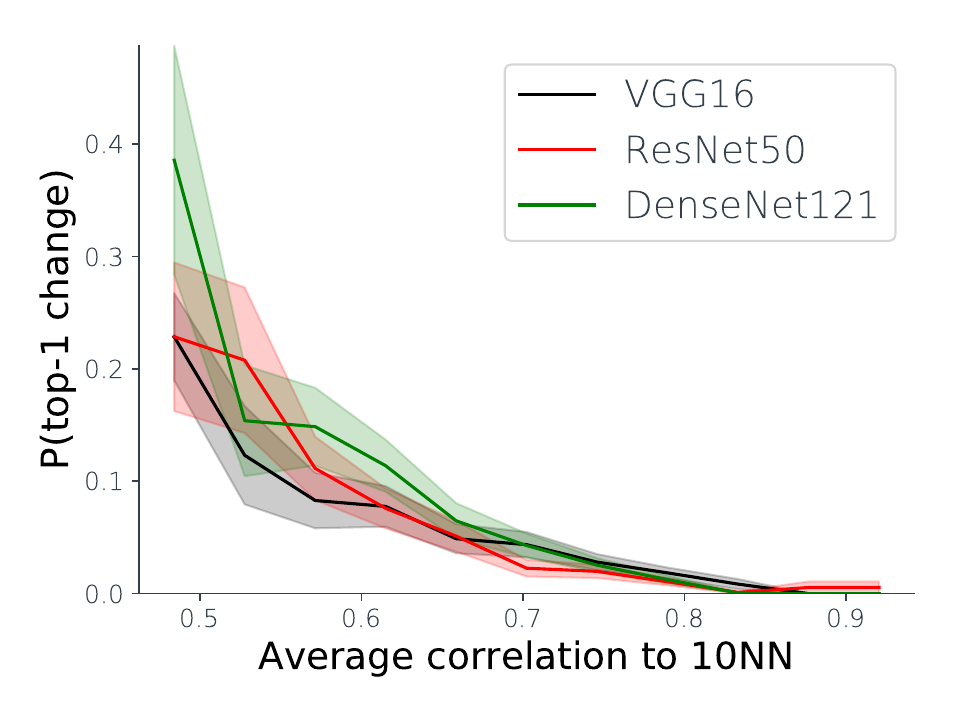}  &
        \includegraphics[width=.4\linewidth]{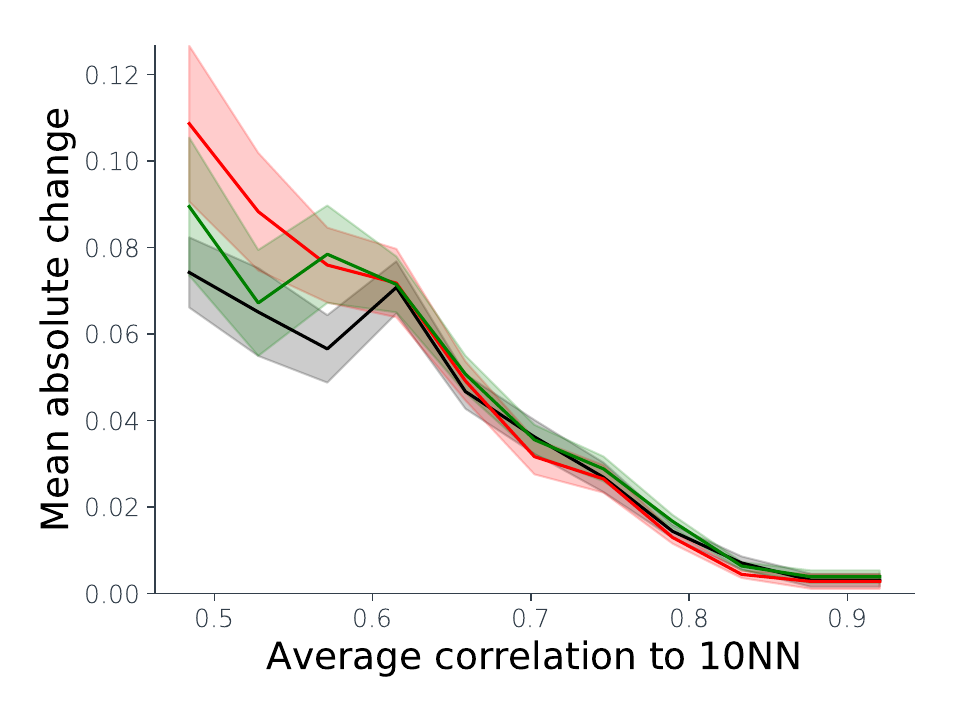} \\[\abovecaptionskip]
  \end{tabular}

  \caption{CNN sensitivity to one pixel perturbations depends on how typical the images are. We plot our two measures of sensitivity as a function of typicality for 1000 randomly chosen, center cropped images. Both measures of sensitivity depend strongly on the typicality of the images: when images obey the bias of photographers, sensitivity is low. Error bars represent the standard error of the mean.}\label{fig:myfig}
\end{figure}

While the preceding discussion suggests that the CNN architecture will
not yield translation invariance ``for free", there is still the
possibility that the CNN will learn a translation invariant prediction
from the training examples, i.e. by learning representations that satisfy the conditions for shiftability. This can be achieved by the network learning filters whose Fourier transform is predominantly concentrated in low frequencies. We would expect the network to learn such filters when they are  trained using data augmentation. As mentioned
previously, the input to the networks were obtained using a random
crop of a training image, so that the network was equally likely to
see a particular image and a one pixel translation of that image. Why
doesn't this cause the network to learn a discriminant function that
is invariant to one pixel translations?

One possible answer to this question is that the network can drive the
training error to zero by learning a discriminant function that is only
invariant to translations of the training images. There is nothing in
the training loss to encourage it to learn a discriminant function
that is invariant to the translation of {\em any input
  image}. Of course we expect the network to generalize what it
learned to images that it did not see during training, but this
generalization may be limited to images that are similar to those seen
during training.

The notion of ``images similar to those seen during training" is particularly relevant if the training set includes significant bias.
Consistent with previous results on ``dataset bias"~\citep{simon2007scene,raguram2008computing,berg2009finding,torralba2011unbiased,weyand2011discovering,mezuman2012learning} we find that the ImageNet dataset is extremely biased in terms of the available sizes and locations of objects. As one illustration, figure~\ref{fig:4}~ shows the distribution of the locations and distances between the
eyes of a ``Tibetan terrier". Note that almost all images of that category show a centered closeup of the dog's face. Thus a network can drive the training error to zero, by being invariant to translations within that narrow range of possible sizes. If this is the case, then we would expect the learned invariance to fail once we present the network with examples of terriers of different sizes. Indeed, the difference in results in figure~\ref{fig:2} is consistent with this view.  When we use the crop protocol, then the object sizes are similar to what was observed during training, but when the embedding protocol is used, the sizes may be very different. Is the higher sensitivity in the embedding protocol due to the fact that this causes
images to be less similar to the training images, or is there
something intrinsic to the embedding protocol that makes networks more
sensitive to perturbations?

Figure~\ref{fig:6} summarizes experiments that suggest that the
important effect is the similarity to the training images. We used the
same protocol as was used in figure~\ref{fig:2}: we randomly chose
images from the ImageNet validation set, resized them, and embedded
them in a larger image. We systematically varied the size of the
embedded images: when the embedded size is close to the input the
network expects (224 for VGG16 and ResNet50,299 for InceptionResNetV2)
then the resizing operation yields images that are similar to those
that the network saw during training. However, when the embedded size
is very different from the size that the network expects, then the
resizing operation yields images that are less similar to the ones
that the network saw during training. If most of the translation
invariance we see in modern CNNs is due to the network learning
partial invariance during its training, then we should expect the
network to be more invariant for large embedded images and less
invariant for small embedded images.

Figure~\ref{fig:6} shows this intuition to be correct. For embedded
sizes close to 220, a one pixel shift or scaling of an image causes a
change in the network prediction approximately 10\% of the time, while
for embedded sizes close to 100, this probability increases to approximately
30\% for the modern networks.

%In summary, whether we use our own hand labeling to measure invariance or we use the available bounding boxes, the ImageNet dataset has strong biases for object position and size and we cannot expect a learning system to learn an invariant representation.

For a more direct test of the hypothesis that the networks only learn
invariance to images that are similar to what it saw during training,
we quantified the sensitivity as a function of the typicality of an input image. Typicality is defined via the ``perceptual similarity'' of an image to the 10 nearest neighbors in the training
set, where``perceptual similarity'' between the two images is the
normalized correlation between the features in the penultimate layer
of ResNet50. Various authors have shown that this definition of perceptual similarity is invariant to the particular network that is used to define the features~\citep{zhang2018unreasonable}. 

Figure~\ref{fig:typicality} shows examples of typical and atypical images for various categories. When an image is taken in a way that satisfies the bias of photographers (e.g. when a dog is captured with a closeup that shows the face in the center of the frame) then the nearest neighbors have a high degree of perceptual similarity. But when when the same dog is photographed at a distance, together with several other objects, the image does not satisfy the bias of photographers and the nearest neighbors have a much lower degree of similarity.

Figure~\ref{fig:myfig} shows that this measure very
clearly predicts the sensitivity of the network to one pixel
perturbations. When a test image has high perceptual similarity to the
training images, then both measures of sensitivity approach zero. But
when an image has low perceptual similarity, the average sensitivity
is quite high (P(top-1 change) approaches 0.3), even though all images here
use the crop protocol that was also used during data augmentation. Note also the similarity between the probabilities of failure of the three different networks: since they were trained using the same biased dataset, all three networks learn to be invariant only to images that satisfy the photographer's bias.

The similarity of an image to its most similar training examples is expected to also change the confidence of the CNN in its prediction of the image: images that have many similar neighbors should yield more confident predictions from the CNN. In the appendix we show that using one CNN's confidence also predicts the amount of invariance other CNNs will exhibit, although the effect seems weaker than when we directly use the similarity to training examples (figure~\ref{fig:confidencevsjagg}).

\section{Possible Solutions}
To summarize, our results show that modern CNNs are not shift invariant and a one pixel shift of an input image can drastically change the network's output. This is despite the convolutional architecture and the use of data augmentation during training. We now discuss three possible solutions and examine the extent to which they solve the problem.

\subsection{Antialiasing}
\label{sec:antialiasing}
As we discussed in section~\ref{sec:Ignoring_the_Sampling_Theorem}, an important principle
of signal processing is that one should always blur before
subsampling, and modern CNN architectures do not obey this
principle. In the recent work of~\citep{zhang2019making}, an elegant proposal
was made regarding how to embody this principle in modern
CNNs. Specifically, they proposed performing the pooling operation in
two steps: a max-pool layer with stride 1, followed by a convolutional
layer with stride=2. Thus the two layers in combination perform
bluring before subsampling. They showed that by changing modern CNNs
to incorporate these two layers instead of the standard max-pool
layers and then retraining, they can significantly improve the
invariance of networks trained on CIFAR10. However, for networks trained on Imagenet, the improvement was much smaller.

\begin{figure}[ht!]
%   \vspace{\floatsep}
\centering
\begin{tabular}{ccccc}
     
%  & ~~~~~~Typical & ~~~~~Atypical \\
 \raisebox{32pt}[0pt][0pt] &
 
 \includegraphics[width=0.11\textwidth]{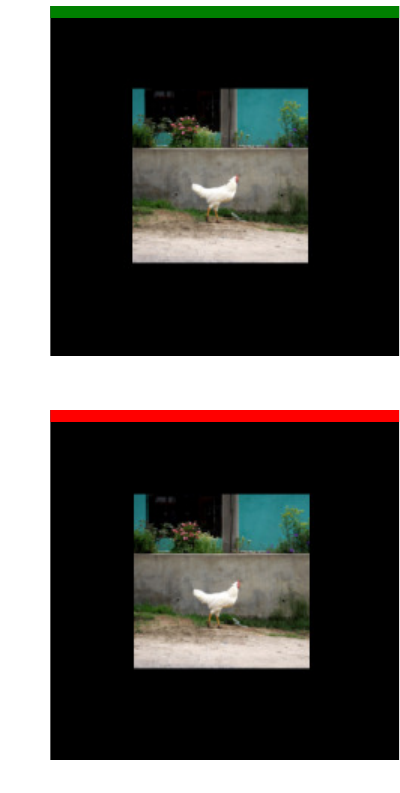} & 
 \includegraphics[width=0.20\textwidth]{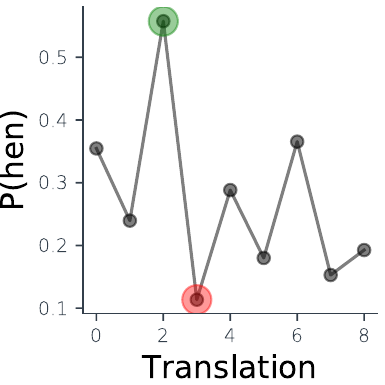} &
 \includegraphics[width=0.11\textwidth]{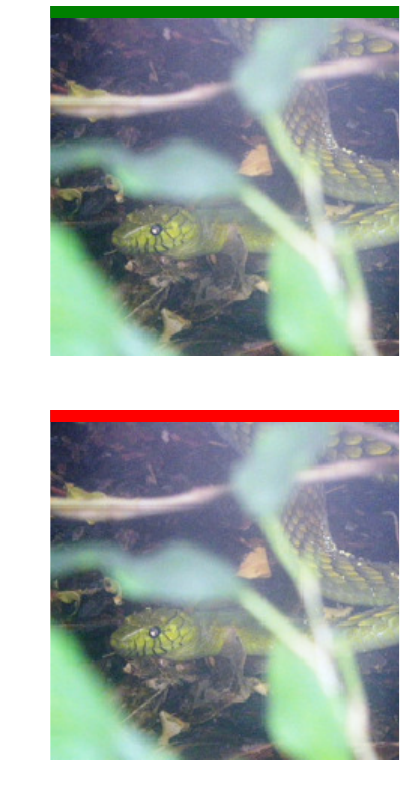} & 
 \includegraphics[width=0.20\textwidth]{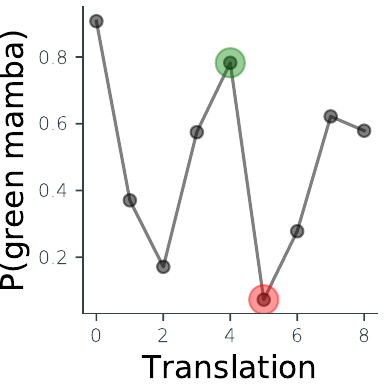} 
%  \raisebox{43pt}[0pt][0pt] &   \\
%  \raisebox{43pt}[0pt][0pt] &   \\
%  \raisebox{43pt}[0pt][0pt] &  
 \end{tabular}
  \begin{tabular}{@{}lr@{}}
    \includegraphics[width=0.68\linewidth]{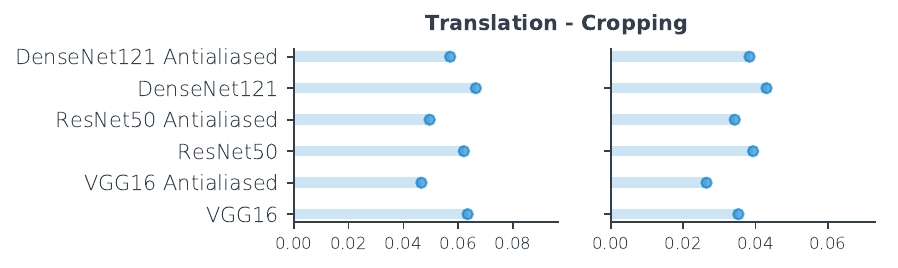} &
    \\[\abovecaptionskip]
    % \small (b) Another image
  \end{tabular}
  
%   \vspace{\floatsep}

  \begin{tabular}{@{}lr@{}}
    \includegraphics[width=0.68\linewidth]{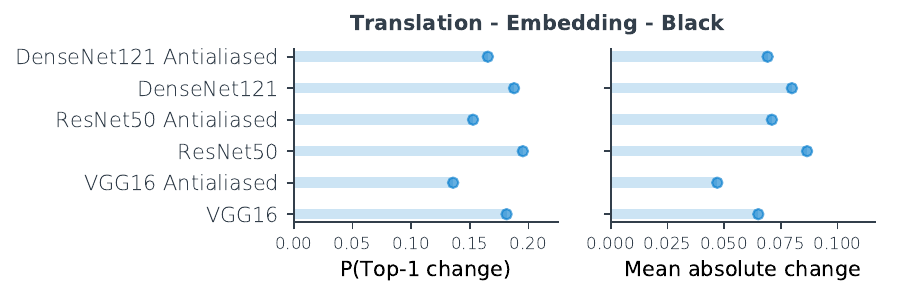} &
    \\[\abovecaptionskip]
    % \small (b) Another image
  \end{tabular}
  
  \begin{tabular}{@{}c@{}@{}c@{}}
    \includegraphics[width=.4\linewidth]{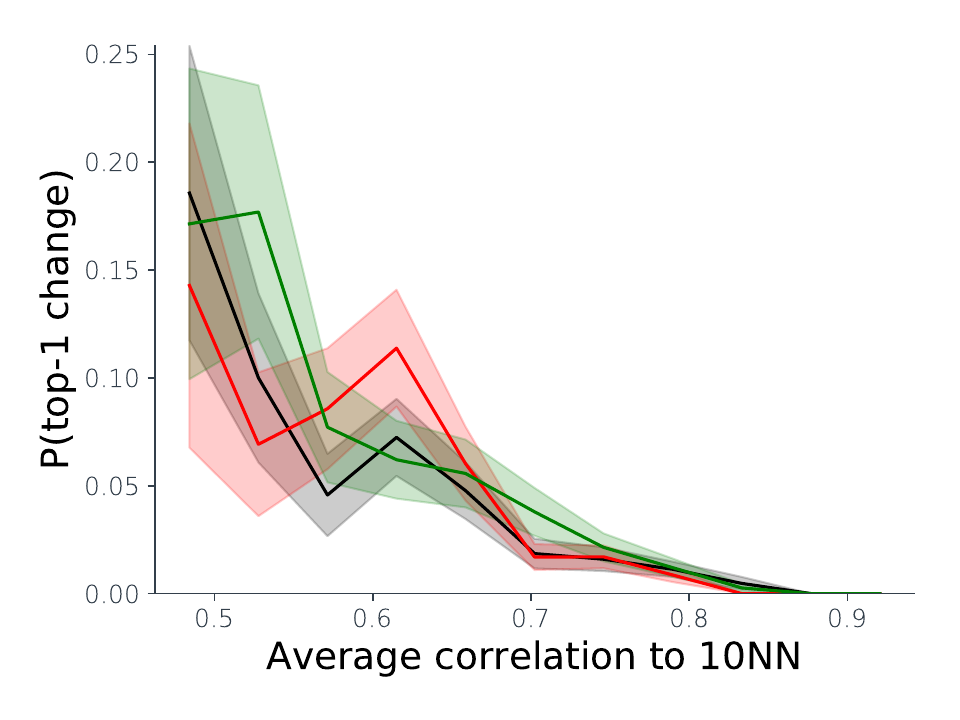}  &
        \includegraphics[width=.4\linewidth]{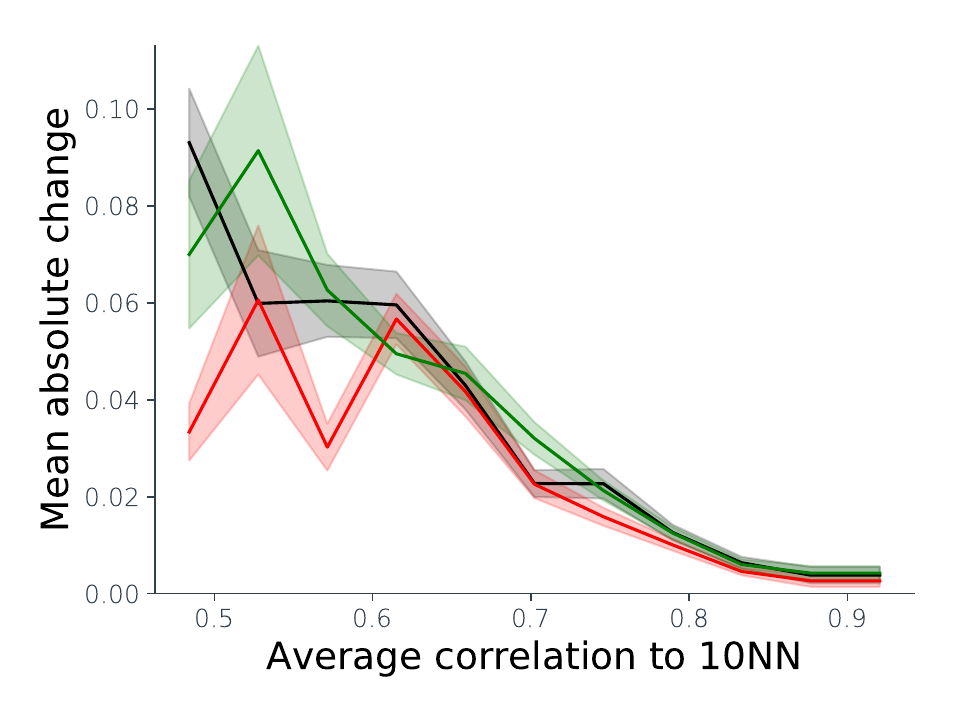} \\[\abovecaptionskip]
  \end{tabular}
  
    \caption{Antialiasing is only a partial solution. Top: Examples of jagged predictions following a 1-pixel translation in the ``Black" and ``Crop" protocols (see figure~\ref{fig:AntialiasingFailuresExamples} for more examples). Middle: a comparison of the senstitivity measures for two of the protocols from figure \ref{fig:2}  with antialiased versions of three modern CNNs (``bin-5" antialiasing taken from \citep{zhang2019making}) and the non-antialiased ones. The bottom shows sensitivity as a function of image typicality when using the crop protocol. When images are not typical, antialiased networks still show high degree of sensitivity to single pixel translations.
    } \label{fig:pytorchAntialiasing}
\end{figure}

Figure~\ref{fig:pytorchAntialiasing} shows that when the antialiased
networks are tested using the protocols that we investigated,
then the effect of antialiasing is relatively small. In particular, when the embedding protocol is used and the size of the embedded image is different from what the network saw during training, a one-pixel shift can still cause the network to change its prediction in about 15\% of the time. Similarly, the bottom of figure~\ref{fig:pytorchAntialiasing} shows the same dependence on typicality in the antialiased networks as we saw in the original networks. When images are similar to those seen during training, then the antialiased networks are invariant, but for atypical images there is still strong sensitivity to one pixel shifts (up to 20\% chance that a one-pixel shift will change the top-1 prediction). Clearly, the antialiased networks do not satisfy our definition of a fully translation-invariant system, which should be invariant to the translation of any pattern.

\begin{table}
    \centering
    \begin{tabular}{c|c|c|c|c}
    & signal1 & signal2 & global sum1 & global sum2\\
    \hline
    Original: & (0,0,0,1,0,0,0) & (0,0,1,0,0,0,0) & 1 & 1\\
    Blurred: & (0,1,4,6,4,1,0) & (1,4,6,4,1,0,0) & 16 & 16 \\
    Sampled: & (0,4,4,0) & (1,6,1,0) & 8 & 8 \\
    ReLU (bias=5): & (0,0,0,0) & (0,6,0,0) & 0 & 6
    \end{tabular}
    \caption[]{When nonlinearities are present, blurring before subsampling does {\em not} guarantee shiftability.}
    \label{table-2}
\end{table}

We believe the reason for
the lack of significant improvement is what we discussed in
section~\ref{sec:Ignoring_the_Sampling_Theorem}: even though blurring before subsampling is
sufficient for avoiding aliasing in linear systems, the presence of
nonlinearities may introduce aliasing even in the presence of blur
before subsampling. Table~\ref{table-2} shows an example: when we use the bin5 filter and a subsampling ratio of 2, a linear system is still shiftable following the subsampling. But this is {\em not} the case when we use a ReLU nonlinearity: despite using the same filter and the same subampling ratio, the system is no longer shiftable. In the appendix, we show a similar example when the bin5 filter is replaced with the ideal low pass filter.  This means that the question of whether the system will be shiftable or not depends on the learnt weights and biases, and we cannot guarantee shiftability simply by virtue of the blurring before subsampling.

%\begin{figure}[ht!]
%\centering
%\centering
%   \begin{tabular}{@{}lr@{}}
%     \includegraphics[width=0.7\linewidth]{Cifar_Accuracies_augmentation.pdf} &
%     % \includegraphics[width=0.3\linewidth]{_Crop_image.pdf}
%     \\[\abovecaptionskip]
%     % \small (b) Another image
%   \end{tabular}
  
% %   \vspace{\floatsep}
\subsection{Increased Data Augmentation}

Engstrom et al.~\citep{engstrom2017rotation} suggested another possible solution for
the lack of translation invariance: increased data augmentation. They
first showed  that networks that were trained using the
standard data augmentation protocol were not invariant when tested with images that
were resized and translated with a black background (the second
protocol in our section~\ref{sec:quantifying_jaggedness}). They then showed
that by training the networks with additional data augmentation that
also included the second protocol, they could increase the
invariance. 

\begin{figure}
\centerline{      
    \includegraphics[width=0.85\linewidth]{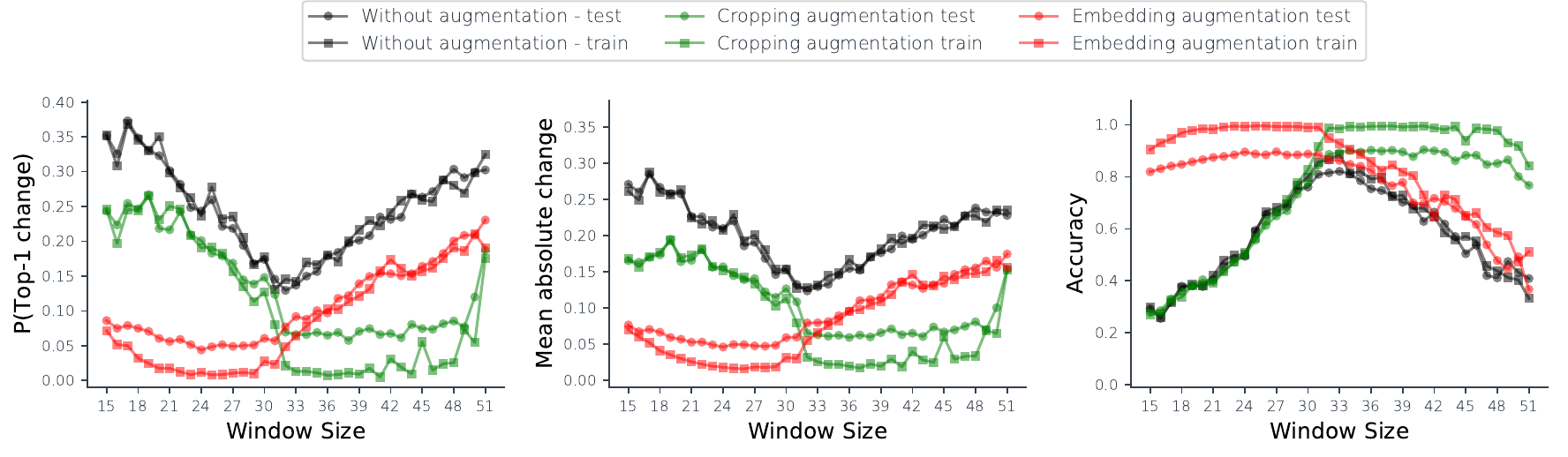} }
    \caption[]{Possible solution - data augmentation. Plotted are two ResNet50 CNNs trained on the CIFAR-10 dataset: Gray - training without data augmentation. green - training with data that is randomly downsized and translated (black background). Red - training with data that is randomly upscaled and cropped. Left - the probability of change to the top-1 prediction following a 1-pixel translation. Middle: the mean absolute change in prediction following a 1-pixel translation. Right - Accuracy of the same networks. Notice that data augmentation is effective mainly in the range of transformations used during training. 
    } \label{fig:CIFAR_Augmentation}
\end{figure}

Figure~\ref{fig:CIFAR_Augmentation} shows experiments we have conducted
using the CIFAR10 dataset that aims to study what CNNs learn from data
augmentation. Do they learn to be invariant to the translation of any
input image, or do they simply learn to be invariant for images that
are visually similar to those seen during training?  The results
support the second alternative: data augmentation mainly causes an
increase in translation invariance for images that were obtained using
the same protocol that was used during training. We do not see any
evidence of the network generalizing from augmented data to the
general notion of translation invariance.  

\subsection{Reducing subsampling}
\begin{figure}
\centerline{
  \includegraphics[width=0.9\linewidth]{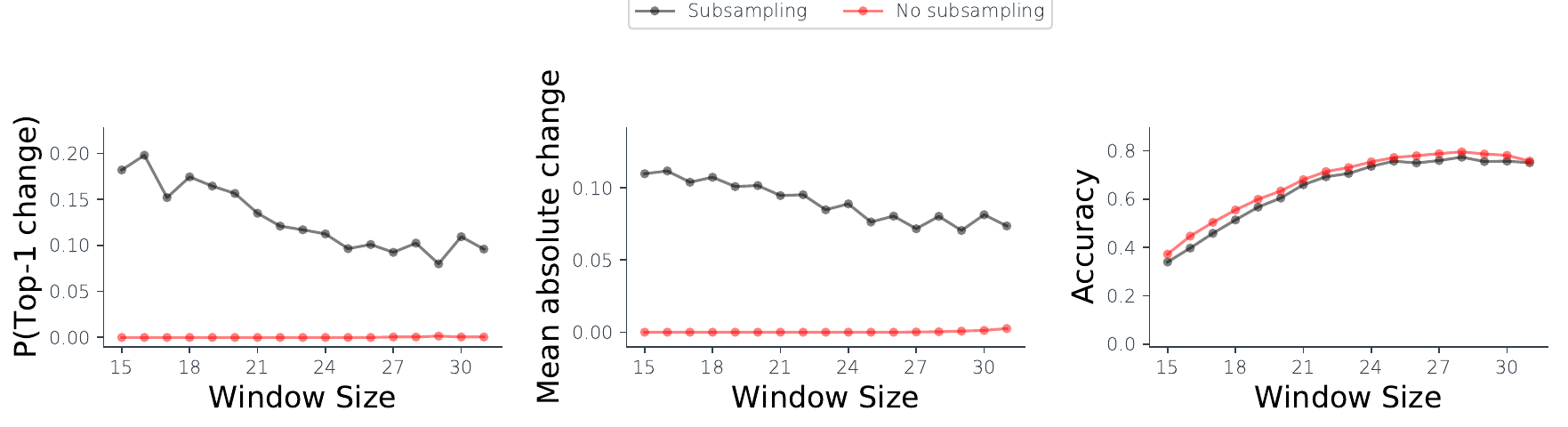}
  }
   \caption{A CNN without subsampling is perfectly translation invariant. Plotted are two Alexnet style CNNs on the CIFAR-10 dataset: with (gray) and without (red) subsampling. The two CNNs achieve similar accuracies of about 0.8.} 
   \label{fig:CIFAR}
\end{figure}
The simplest way to reduce aliasing artifacts is to avoid subsampling. Unfortunately this comes with a large computational price but for small images this is possible. Figure~\ref{fig:CIFAR} shows that when we train a subsampling-free CNN on CIFAR10 images it achieves high accuracy and is now truly translation invariant. This does not seem feasible for larger size images (although see~\citep{chen2014semantic,chen2017deeplab,yu2015multi,yu2017dilated} for ways to make this more efficient). In these experiments, we tested the network with images that were rescaled and embedded into a black image, and the network never saw images of this size during training. Nevertheless, the convolutional architecture without subsampling guarantees translation invariance for any input pattern.

\section{Discussion}

CNN architectures were designed based on an intuition that the
convolutional structure and pooling operations will give invariance to
translations and small image deformations ``for free''. In this paper
we have shown that this intuition breaks down once subsampling, or
``stride'' is used and we have presented empirical evidence that
modern CNNs do not display the desired invariances since the
architecture ignores the classic sampling theorem. This still leaves
open the possibility of a CNN learning invariance from the data
augmentation procedure  but we have shown that this is not the
case. Rather, data augmentation teaches the network to be invariant to
translations but only for images that are visually similar to typical
images seen during training, i.e. images that obey the photographer's
bias. 

In addition to pointing out these failures, the sampling theorem also suggests a way to impose translation invariance by ensuring that all representations are sufficiently blurred to overcome the subsampling.  Alternatively, one could use specially designed features in which invariance is hard coded %\citepp{chomat2000local,lindeberg1994scale,lowe1999object,lowe2004distinctive} 
or neural network architectures that explicitly enforce invariance \citep{sifre2013rotation,gens2014deep,cheng2016learning,cheng2016rifd,dieleman2016exploiting,dieleman2015rotation,xu2014scale,worrall2017harmonic,cohen2016group}. However, trying to guarantee invariance to transformation for {\em any} input pattern, may end up hurting performance in datasets that contain significant photographer's bias.

\acks{We thank Tal Arkushin for the helpful comments. Support by the ISF and the Gatsby Foundation is gratefully acknowledged.}

\bibliographystyle{icml2019}
\bibliography{main.bib}

% Acknowledgements should go at the end, before appendices and references

% Manual newpage inserted to improve layout of sample file - not
% needed in general before appendices/bibliography.

\newpage

\appendix
\section{}

\subsection{Pipeline for producing the bottom row of figure 1}
We download this video: https://youtu.be/akseo5DuXgU using an online downloader. We load the video frames and crop them to a 1x1 aspect ratio while making sure that the object is visible for the entire duration of the video. After the crop, we resize the frames to 299 by 299 as used by the standard Keras applications framework (https://keras.io/applications/). We preprocess the frames using the standard Keras preprocessing function. Finally, we use the predictions of the InceptionResNetV2 model to demonstrate the jagged behavior shown in figure 1.

\subsection{The network architectures used in figure \ref{fig:2}}
\begin{table}[ht!]
\begin{center}
        \begin{tabular}{lcccc}\hline
        Network & Top-1 & Top-5 & Parameters & Depth \\
        \hline
        VGG16 & 0.715 & 0.901 & 138,357,544 & 16 \\
        ResNet50 & 0.759 & 0.929 & 25,636,712 & 50\\
        InceptionResNetV2 & 0.804 & 0.953 & 55,873,736 & 134\\ \hline
        \end{tabular}
        \caption{The networks used (taken from (\url{https://keras.io/applications/}))}\label{tab:a}
        \label{table:NETWORKS}
    \end{center}
% \caption{My table}

\end{table}

%\subsection{Measuring the recognizability of embedded images}
%We asked three human subjects to recognize images taken from the %ImageNet dataset after they were resized to size 100x100 and embedded %within a large image of size 224x224. Following embedding, we used the %same inpainting procedure as before. The average recognition accuracy %was 93\% for the embedded images and average recognition accuracy for %the original images was 94\%.

\subsection{ReLU causes aliasing even with ideal low pass filter}
Table~\ref{table-2} shows an example of how nonlinearity can destroy shiftability even if blurring is used prior to subsampling. This example can be extended to the case when the blur filter is the ideal low pass filter. In this case, the two blurred signals will simply be infinite signals $sinc(k/2),sinc((k-1)/2)$, whose central part is $(0,0.6336,1,0.6336,0)$. After subsampling, the first signal will be a delta function, whose central part is $(0,0,1,0,0,0)$ and the shifted signal will be infinite with a central part of $(-0.2122,0.6336,0.6336,-0.2122)$. The global sum of both these signals is exactly 1, showing again that shiftability preserves the global sum. But if we apply a ReLU with any bias between $0.6336$ and $1$, then the first signal will remain the delta function, while the second signal will be zero everywhere.

\subsection{CNN confidence can predict sensitivity}

\begin{figure}
  \centering
  \begin{tabular}{@{}c@{}@{}c@{}}
    \includegraphics[width=.4\linewidth]{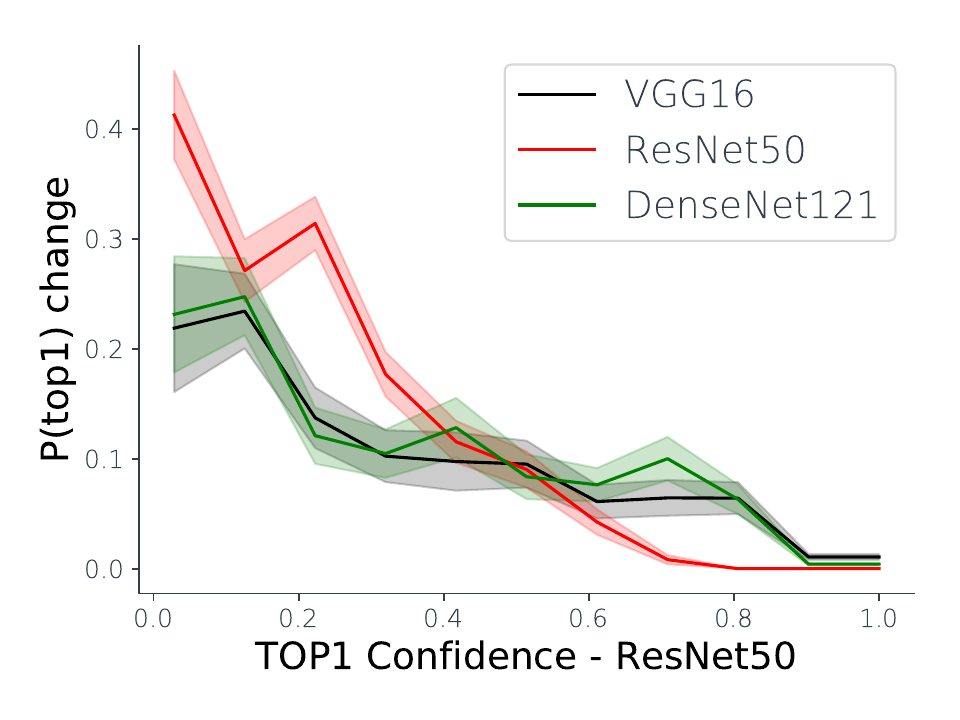}  &
        \includegraphics[width=.4\linewidth]{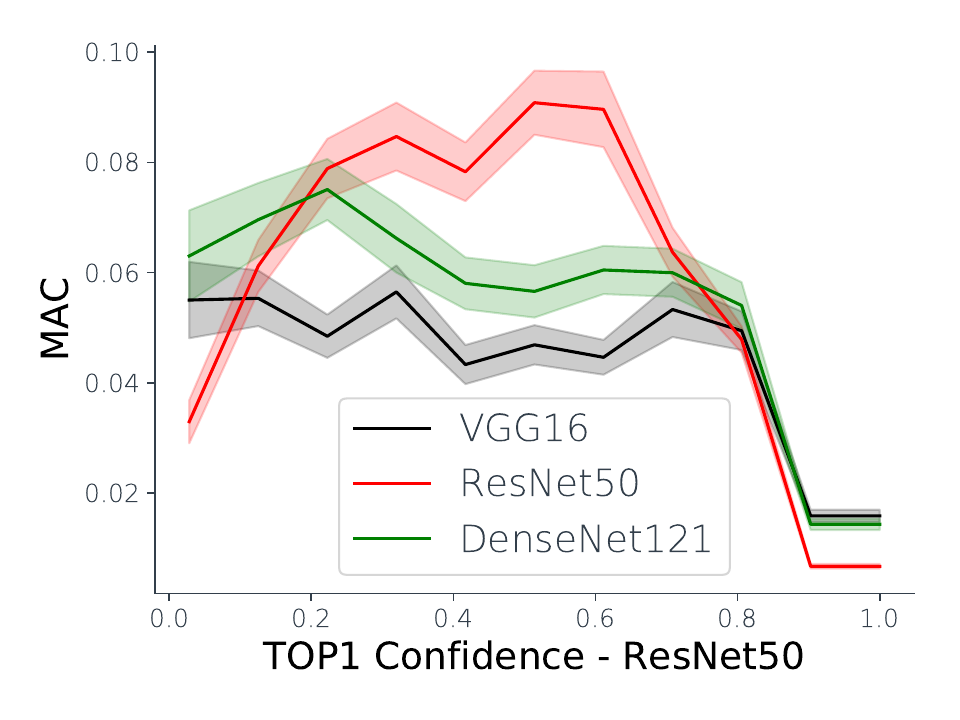} \\[\abovecaptionskip]
  \end{tabular}

  \caption{CNN sensitivity to one pixel perturbations depends on the confidence the network assigns to the top1 prediction. We plot our two measures of sensitivity as a function of top1 confidence for 1000 randomly chosen, center cropped images. Error bars represent the standard error of the mean.}\label{fig:confidencevsjagg}
\end{figure}

We have shown that the similarity of an image to its 10 most similar training examples can predict the sensitivity to a one pixel perturbation by different CNNs. We also expect the CNN's confidence to vary as a function of this similarity: when an image has many very similar training images, we expect the CNN's confidence in its prediction to be large. This suggests using the CNN's confidence for a given image as a predictor of how sensitive different networks will be to one pixel perturbations of that image. Figure~\ref{fig:confidencevsjagg} shows that this can be used as a predictor, but the effect seems to be weaker than using the similarity of an image to its most similar training examples.

\subsection{More examples of failure of antialiased networks}
\begin{figure}[ht!]
%   \vspace{\floatsep}
\centering
\begin{tabular}{ccccc}
     
%  & ~~~~~~Typical & ~~~~~Atypical \\
%  \raisebox{32pt}[0pt][0pt] &
 
%  \includegraphics[width=0.30\textwidth]{AntialiasedFailures/BlackExampleImages_5.pdf} & 
%  \includegraphics[width=0.15\textwidth]{AntialiasedFailures/BlackExample_5.pdf} &
%  \includegraphics[width=0.30\textwidth]{AntialiasedFailures/CropExampleImages_4724.pdf} & 
%  \includegraphics[width=0.15\textwidth]{AntialiasedFailures/CropExample_4724.pdf} \\
%  \raisebox{43pt}[0pt][0pt] &  \includegraphics[width=0.30\textwidth]{AntialiasedFailures/BlackExampleImages_6277.pdf} & 
%  \includegraphics[width=0.15\textwidth]{AntialiasedFailures/BlackExample_6277.pdf} & \includegraphics[width=0.30\textwidth]{AntialiasedFailures/CropExampleImages_5417.pdf} & 
%  \includegraphics[width=0.15\textwidth]{AntialiasedFailures/CropExample_5417.pdf}
%  \\
 \raisebox{43pt}[0pt][0pt] &  \includegraphics[width=0.30\textwidth]{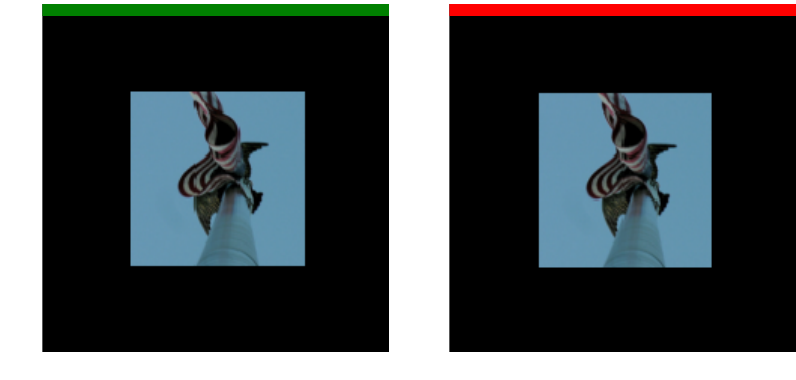} & 
 \includegraphics[width=0.15\textwidth]{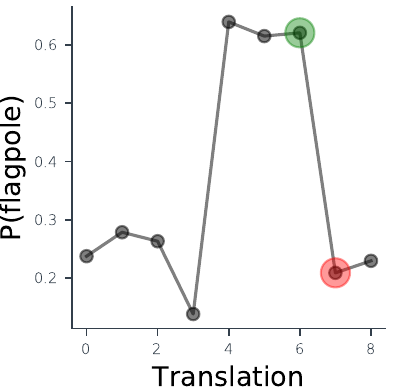} &
 \includegraphics[width=0.30\textwidth]{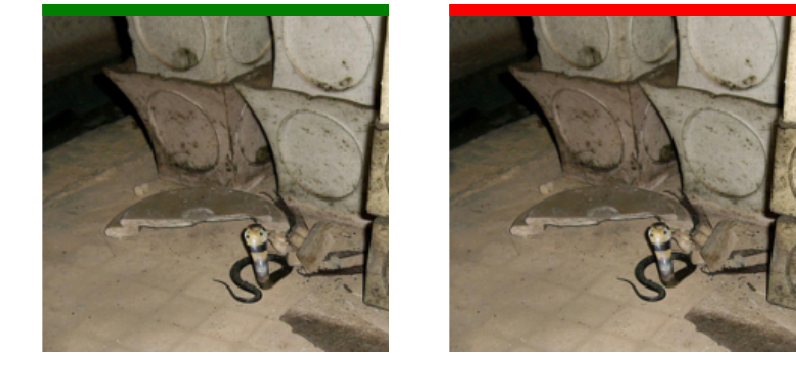} & 
 \includegraphics[width=0.15\textwidth]{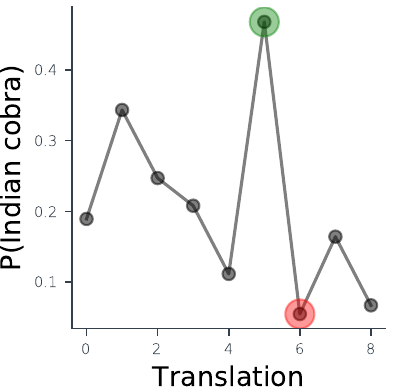}
 \\
 
  \raisebox{32pt}[0pt][0pt] &
 
 \includegraphics[width=0.30\textwidth]{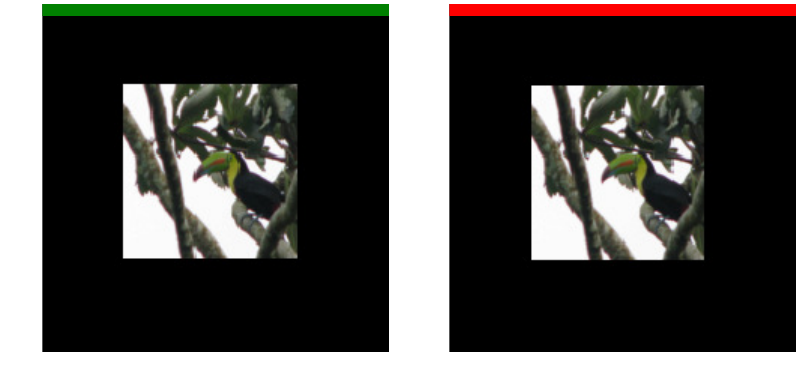} & 
 \includegraphics[width=0.15\textwidth]{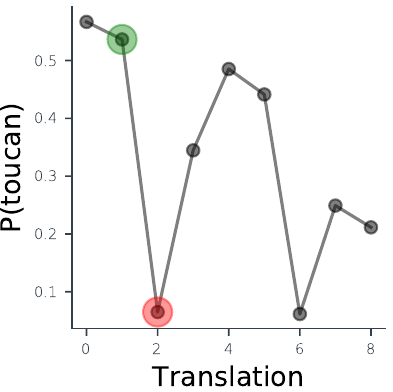} &
 \includegraphics[width=0.30\textwidth]{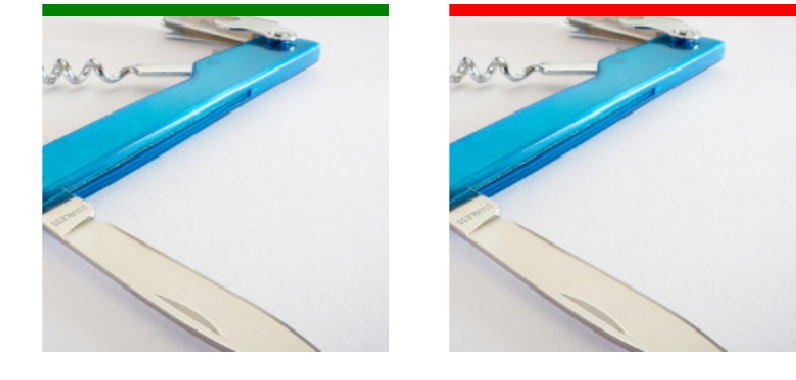} & 
 \includegraphics[width=0.15\textwidth]{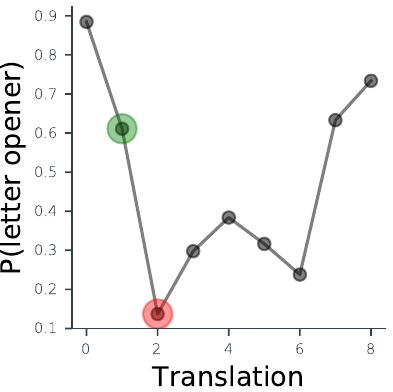} \\
 \raisebox{43pt}[0pt][0pt] &  \includegraphics[width=0.30\textwidth]{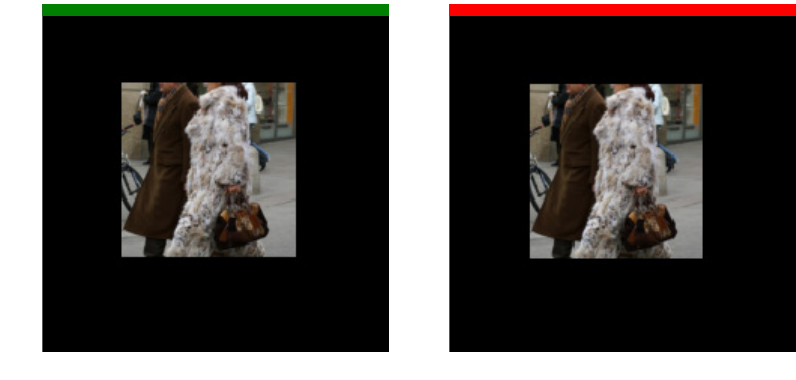} & 
 \includegraphics[width=0.15\textwidth]{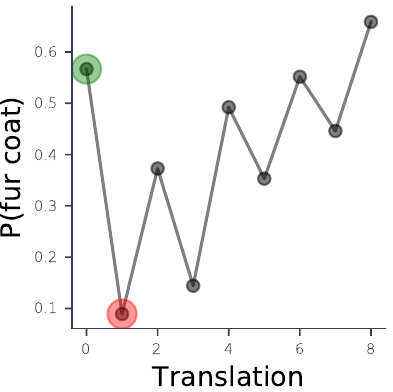} & \includegraphics[width=0.30\textwidth]{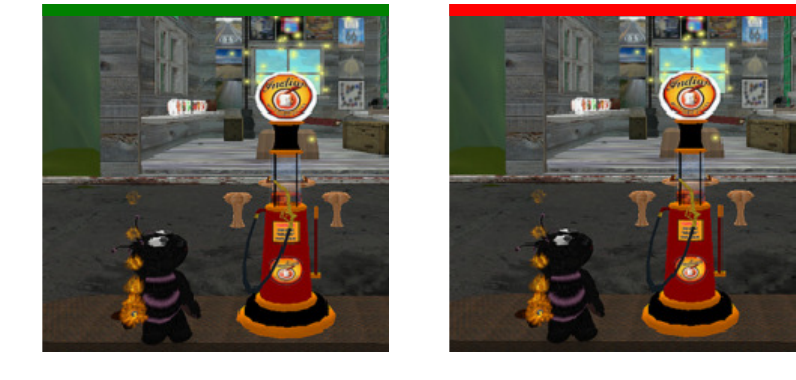} & 
 \includegraphics[width=0.15\textwidth]{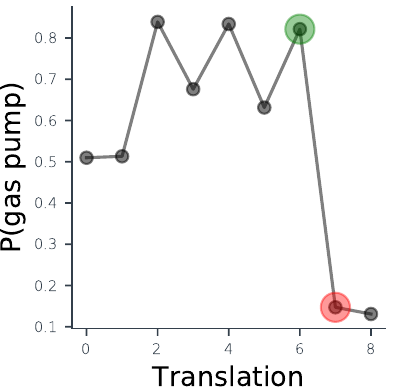}

%  \\
%  \raisebox{43pt}[0pt][0pt] &  \includegraphics[width=0.30\textwidth]{AntialiasedFailures/BlackExampleImages_8768C.pdf} & 
%  \includegraphics[width=0.15\textwidth]{AntialiasedFailures/BlackExample_8768C.pdf} &
%  \includegraphics[width=0.30\textwidth]{AntialiasedFailures/CropExampleImages_1741C.pdf} & 
%  \includegraphics[width=0.15\textwidth]{AntialiasedFailures/CropExample_1741C.pdf}
 \\
%  \raisebox{43pt}[0pt][0pt] &   \\
%  \raisebox{43pt}[0pt][0pt] &   \\
%  \raisebox{43pt}[0pt][0pt] &  
 \end{tabular}
     \caption[]{Additional examples of failures of antialiased networks in the ``Black" and ``Crop" protocols.} 
     \label{fig:AntialiasingFailuresExamples}
\end{figure}

\newpage

\vskip 0.2in

\end{document}